\documentclass[a4paper,12pt]{article}

\usepackage{amsmath,amsfonts,amssymb,amsthm}

\usepackage[polish,british]{babel}

\usepackage{url}

\usepackage[T1]{fontenc}
\usepackage[utf8]{inputenc}
\usepackage{lmodern}

\usepackage{comment}  %

\usepackage{indentfirst}
\usepackage{graphicx}
\usepackage{booktabs}    %
\usepackage{accents}    %
\usepackage{url}
\usepackage{siunitx}

\usepackage{color}

\usepackage{lineno} %

\usepackage{hyperref}
\usepackage{tikz}
\usetikzlibrary{calc,arrows}
\usepackage{filecontents} 
\usepackage{pgfplots}
\usepackage{pgfplotstable}

\newcommand{\orcid}{0000-0002-6118-606X} %
\newcommand{\ksorcid}{\orcid}

\newcommand{\Membership}{u}

\iflanguage{polish}{%

}%
{%

}

\DeclareMathOperator*{\tnorm}{\star}

\usepackage{thmtools}                    %

\declaretheorem[style=definition,qed=$\blacksquare$,numberwithin=section,name=Definition]{definition}
\declaretheorem[style=definition,qed=$\blacktriangle$,sibling=definition,name=Example]{example}
\declaretheorem[style=definition,qed=$\lozenge$,sibling=definition,name=Remark]{remark}

\newtheorem{theorem}{Theorem}

\usepackage{listings}
\newcommand{\lS}{<_S}
\newcommand{\eqS}{=_S}
\newcommand{\gS}{>_S}
\newcommand{\leqS}{\leqslant_S} 
\newcommand{\geqS}{\geqslant_S}

\newcommand{\ksR}[3]{#1\left(#2,#3\right)}

\title{\sffamily\bfseries Fuzzy numbers revisited: operations on extensional fuzzy numbers}
\author{Krzysztof Siminski\\Silesian University of Technology, Gliwice, Poland\\\url{krzysztof.siminski@polsl.pl}\\\ksorcid}

\begin{document}
\maketitle

\begin{abstract}
Fuzzy numbers are commonly represented with fuzzy sets.
Their objective is to better represent imprecise data.
However, operations on fuzzy numbers are not as straightforward as maths on crisp numbers.
Commonly, the Zadeh's extension rule is applied to elaborate a result.
This can produce two problems:
(1) high computational complexity and 
(2) for some fuzzy sets and some operations the results is not a fuzzy set with the same features (eg. multiplication of two triangular fuzzy sets does not produce a triangular fuzzy set).
One more problem is the fuzzy spread – fuzziness of the result increases with the number of operations.
These facts can severely limit the application field of fuzzy numbers.
In this paper we would like to revisite this problem with a different kind of fuzzy numbers – extensional fuzzy numbers.
The paper defines operations on extensional fuzzy numbers and relational operators ($=$, $>$, $\geqslant$, $<$, $\leqslant$) for them.
The proposed approach is illustrated with several applicational examples.
The C++ implementation is available from a public GitHub repository.
\end{abstract}

\newcommand{\efn}{extensional fuzzy number}
\newcommand{\Efn}{Extensional fuzzy number}

\newcommand{\OWA}{\mathrm{OWA}}
\newcommand{\XS}{\X^*}
\newcommand{\wXS}{\wX^*}

\section{Introduction}

Fuzzy numbers are commonly represented with fuzzy sets. 
Their objective is to better handle imprecision of data.
There are multiple representations of fuzzy numbers, eg. interval, triangular fuzzy sets. 
Linguistic interpretability of fuzzy numbers is their huge advantage. 
They can be used to represents linguistic terms (eg. “more or less 7”, “almost 4”, “slightly more than 5”). 
Description of numbers with linguistic terms are very common in human. 
In everyday practice, we use linguistic numbers more frequently than precise numbers. 
One more fact is very important. Human not only understand linguistic terms for numbers, but can also operate on them. 
We can without any effort sum up imprecise linguistic numbers (eg. circa 5 and more or less 4). 

Linguistic numbers can be easily represented with fuzzy numbers. 
But operations on fuzzy numbers are not as straightforward as maths on crisp numbers. 
Commonly, the Zadeh's extension rule is applied to elaborate a result.
However application of this rule can produce two problems:
(1) high computational complexity: for some fuzzy sets numerical integration is needed and some frequently used functions do not have closed form formulae for integrals what results in a need for numerical integration; 
(2) for some fuzzy sets and some operations the results is not a fuzzy set with the same features (eg. multiplication of two triangular fuzzy sets does not produce a triangular fuzzy set). 
These two fact can severely limit application field of fuzzy numbers. 

In this paper we would like to revisited this problem with a different kind of fuzzy numbers – \efn s.
In the paper, we discuss construction and properties of \efn s.
Then, we focus on relational operators ($=$, $>$, $\geqslant$, $<$, $\leqslant$) for \efn s.
Finally, we provide some illustrative examples of applications.

\section{Related research}

\subsection{Granular computing}

Lofti Zadeh coined a new paradigm for granular computing in his seminal paper in 1979 \cite{id:Zadeh1979Fuzzy}. After a very long hibernation, this approach came back to life and made a huge progress \cite{id:Yao2013Granular}. Today, when we have many techniques, methods, paradigms, algorithms, and experience \cite{id:Salehi2015Systematic}, granular computing (GrC) is claimed a new emerging field of research in data mining and machine learning \cite{id:Yao2007Art}. 

The main idea of granular computing is to provide ideas, techniques, and method to make computer work in a way similar to human cognition \cite{id:Zadeh1997Toward}. This comprises granulation (decomposition of a whole into parts with preserved semantics), organisation (of parts into a whole), and causation (cause-effect relation). The idea of Zadeh was to make computer calculate with words \cite{id:Zadeh2002Computing,id:Piegat2015Computing}. GrC is also a starting point for human-centric data approach \cite{id:Pedrycz2015Data}. 

To make application of GrC easier, Yao \cite{id:Yao2016triarchic} proposed a three-way approach to granular computing. His triarchic theory of GrC is based on three perspectives, each supporting the other two. First, the philosophical perspective handles the structured thinking with the meronym-holonym concept (part-whole relation). Second, the methodological perspective provides the structured problem solving with methods, algorithms, techniques. Third, the computational perspective focuses on the information processing. 

Granular computing needs granules. Thus the first step is granulation of provided data. The keep the semantics of elaborated granules, the principle of justifiable granulation is applied \cite{id:Pedrycz2013Building}. We can handle this step quite well. The next step is the computing with granules. This is still a huge challenge for researchers. In this paper we would like to propose an idea and technique for efficient operations of fuzzy granules. 

\subsubsection{Information granules}
\label{id:sec:granules}
Information granule is a collection of entities in the sense of similarity, likeness, proximity, indiscernibility, identity, or adjacency \cite{id:Pedrycz2013Granular,id:Yao2007Granular,id:Yao2008GranularPast,id:Siminski2022Prototype,id:Shifei2010Research,id:Siminski2021Outlier,id:Suchy2023GrDBSCAN}. 	
An important feature of granules is their semantics. They can be labelled with semantically rich labels. This makes them a good representation of linguistic terms that are commonly used by humans. 
The second feature of granules is also human-centric: it is a hierarchy of granules. A granules can be split into subgranules when needed. Granules can merge into a higher level supergranules \cite{id:Wang2021Designing,id:Yao2018Three,id:Ciucci2016Orthopairs}.

\subsubsection{Representation of granules}

Common representations are intervals, sets, fuzzy sets, interval type-2 fuzzy sets \cite{id:Pedrycz2012GranularRepresentation}, rough sets \cite{id:Skowron2016Interactive,id:Pieta2021Applications}, intuitionistic sets, shadowed sets, clusters  \cite{id:Siminski2020GrFCM}, soft sets \cite{id:Xia2022New}, and if-then rules \cite{id:Siminski2021GrNFS}.  
A set is a simple representation of a collection of related items. Fuzzy sets enable partial membership, whereas rough sets address a case when no ``yes-no decision'' can be made.  They are a starting point for the three-way decision approach \cite{id:Yao2009Three,id:Yao2011Superiority,id:Siminski20233WDNFS}. 

The representation of granules is heavily combined with the granulation process. For example, granules represented with fuzzy sets may be elaborated with fuzzy clustering. Granules in form of intervals are commonly produces with discretisation of the task domain. Typically, the techniques applied to elaborate granules are discretization, aggregation, transformation \cite{id:Yao2020Three-way}.

\subsection{Fuzzy numbers}

Fuzzy sets are a very common representation of information granules. 
Granular computing (GrC) aims at calculation with granules. 
This is why fuzzy numbers are used in GrC. 
If a fuzzy granule can be treated as a fuzzy number and we know how to operate on fuzzy numbers, we can calculate with fuzzy granules.

Below we present a common definition of fuzzy numbers.
\begin{definition}[\cite{id:Dubois1978Operations}]
A fuzzy number is a fuzzy set $A$ that satisfies the following:
\begin{itemize}
\item $A$ is normalised, ie. $\exists x : \Membership_A(x) = 1$,
\item $A$ is fuzzy convex, ie. $\forall t \in [0,1] : \forall x,y \in R : \Membership_A(tx + (1-t)y) \geqslant \min\{\Membership_A(x), \Membership_A(y)\}$,
\item Support of $A$ is bounded.
\item The membership function of $A$ is piecewise continuous.
\end{itemize}
\end{definition}

This is not the only definition of fuzzy numbers. The paper \cite{id:Ma1999New} provides two more definitions. One of them is:
\begin{definition}[\cite{id:Ma1999New}]
A fuzzy number is a fuzzy set $A$ that satisfies the following:
\begin{itemize}
\item Its membership function $\Membership_A$ is upper semicontinuous;
\item $\Membership_A(x) = 0$ outside some interval $[c,d]$;
\item There are real numbers $a, b : c\leqslant a \leqslant b \leqslant d$ for which
  \begin{itemize}
\item $\Membership_A$ is monotonic increasing on $[c, a]$;
\item $\Membership_A$ is monotonic decreasing on $[b, d]$;
\item $\Membership_A(x) = 1$ for $a\leqslant x \leqslant b$.
\end{itemize}
\end{itemize}
\end{definition}

These definitions are very general and allow for very various construction of fuzzy numbers.

\subsubsection{Representation of fuzzy numbers}

In this section we focus on operations on fuzzy numbers.
They are defined with the the Zadeh's extension principle.
The operation $\ast$ on fuzzy numbers results in \cite{id:Nguyen1978Note}:
\begin{align}
C(z) = A(x) \circledast B(y) = \sup_{z = x \ast y} \min\left(A(x), B(y)\right).
\end{align}

\paragraph{Intervals}
A very simple example of a fuzzy number is an interval. An interval fuzzy numbers is just an interval $[l, r]$. 
Operation $\circledast$ on an interval is defined as \cite{id:Shaocheng1994Interval}:
\begin{align}
\left[l_1, r_1\right] \circledast \left[l_2, r_2\right] = \left\{g_1 \circledast g_2 : l_1 \leqslant g_1 \leqslant r_1, l_22 \leqslant g_2 \leqslant r_2 \right\}.
\end{align}
This definition is based on the Zadeh's extension principle. 
In general, this definition requires discretisation and iterative elaboration of approximate results. However, for addition, subtraction, multiplication, and division closed form formulae can be proved.

\paragraph{Triangular fuzzy numbers}

\begin{definition}
A triangular fuzzy number is defined as a fuzzy set defined with three values $l \leqslant c \leqslant r$, where $(l, r)$ is the support of the set and $c$ is its core. 
\end{definition}

The general approach to operations on fuzzy numbers can be presented in the following steps:
\begin{itemize}
\item discretisation into a finite number of $\alpha$-cuts,
\item operations on $\alpha$-cuts,
\item merging of $\alpha$-cuts into a final result.
\end{itemize}

\begin{definition}
The $\alpha$-cut of a fuzzy set $A$ is defined as a crips set of all element whose membership to the $A$ set is not less than $\alpha$:
\begin{align}
A_\alpha = \left\{  x \in X : \Membership_A (x) \geqslant \alpha \right\}.
\end{align}
\end{definition}

A sum of two triangular fuzzy numbers $A = (l_a, c_a, r_a)$ and $B = (l_b, c_b, r_b)$ is a triangular fuzzy number $A + B = (l_a + l_b, c_a + c_b, r_a + r_b)$ \cite{id:Cheng2004Group}.

Scalar multiplication of a triangular fuzzy number $A = (l_a, c_a, r_a)$ for a crisp multiplier $k\geqslant 0$ is a triangular fuzzy number $kA = (kl_a, kc_a, kr_a)$ \cite{id:Cheng2004Group}.

However, the set of triangular fuzzy sets is not closed under multiplication and division. The paper \cite{id:Mukherjee2023Brief} presents a detailed analysis of operations on triangular fuzzy numbers. They prove that a product and quotient of triangular fuzzy numbers are not triangular fuzzy numbers. Thus triangular fuzzy number are not closed under multiplication and division.  %
This is a problem when we want to execute multiple operations in triangular fuzzy numbers (eg. in a loop).

\paragraph{Gaussian fuzzy numbers}
Application of the Zadeh's extension principle to operations on Gaussian fuzzy sets is decisively more complicated task than for triangular fuzzy sets. There are very few articles on this issue.
The paper \cite{id:Tolga2020Finite} presents some arithmetic on Gaussian fuzzy numbers but without any mathematical background.

\subsection{Disadvantages of classical fuzzy numbers}
Application of the Zadeh's extension principle causes three cardinal problems:
\begin{itemize}
\item computational complexity (in general, discretisation and iterative approach is needed),
\item the results do not preserve the shape of operands (it makes elaboration of results in loops difficult or even impossible),
\item accumulation of fuzziness (the more operations are executed, the more fuzzy the result) \cite{id:GeramiSeresht2019Computational}.
\end{itemize}
These three problems make application of fuzzy numbers based on Zadeh's extension principle hardly acceptable in practice.
This is why in this paper we would like to focus on a different approach to fuzzy numbers.

\section{\Efn s}
This approach is based on a different approach than Zadeh's extension principle. 
The main idea here is an extension of a crisp number with a T-norm-similarity relation.
Here, we only provide the essentials for understanding of extensional fuzzy numbers. 
For details and mathematical background please refer to
\cite{id:Holcapek2012Arithmethics-I,id:Holcapek2012Arithmethics-II,id:Holcapek2014MI-Algebras,id:Stepnicka2020Properties}.

\begin{definition}
A fuzzy relation $R$ between $X$ and $Y$ is a fuzzy set
\begin{align}
\ksR{R}{x}{y} = \left\{\Membership (x, y) \big/ (x,y) : x \in X \land y \in Y\right\},	
\end{align}
where $\Membership$ stands for the membership function.
\end{definition}

Fuzzy $\tnorm$-similarity relation is a crucial notion for \efn s and is defined as follows:
\begin{definition}[\cite{id:Holcapek2012Arithmethics-I}]
\label{def:rel:tnorm-similarity}
A fuzzy $\tnorm$-similarity relation defined for a left continuous t-norm $\tnorm$ satisfies the following properties for all $x, y, z \in \mathbb{R}$:
\begin{align}
\ksR{S}{x}{x} & = 1, \label{eq:S:reflexivity}\\
\ksR{S}{x}{y} & = \ksR{S}{y}{x}, \label{eq:S:symmetry}\\
\ksR{S}{x}{y} \tnorm \ksR{S}{y}{z} &\leqslant \ksR{S}{x}{z}. \label{eq:S:trojkat}
\end{align}
\end{definition}
It is used to construct \efn s.
\begin{definition}[\cite{id:Holcapek2012Arithmethics-I}]
The \efn s are constructed with extension of crisp numbers with a fuzzy $\tnorm$-similarity relation defined for a left continuous t-norm $\tnorm$.
It can be proved that \efn\ $x_S$ of $x$ with regard to $\tnorm$-similarity relation $S$ on $\mathbb{R}$ can be expressed as \cite{id:Holcapek2012Arithmethics-I}:
\begin{align}
\forall_{y \in \mathbb{R}} \quad x_S(y) = \ksR{S}{x}{y}.
\end{align}
\end{definition}

\Efn s have several advantages \cite{id:Holcapek2012Arithmethics-I}:
\begin{itemize}
\item Multiple operations of fuzzy numbers do not increase the vagueness of fuzzy numbers.
\item The result of an operation on \efn s is a \efn.
\item It is possible to use fuzzy numbers with an unbounded support (eg. for the Gaussian fuzzy sets).
\item The computational overload of \efn s is very low: no discretisation, no integration, no loops are needed.
\item A plethora of multiple types of \efn s can be defined with fuzzy $\tnorm$-similarity relations.
\end{itemize}

\subsection{Examples}
\begin{figure}
\centering
\begin{tikzpicture}
\draw[<->] (0,2.5) -- (0,0) -- (6,0);

\draw[dotted] (0,2) node[anchor=east] {$1$} -- +(5,0);

\draw[dashed] (3,0) node[anchor=north] {$x$} -- +(0,2);
\draw (1,0) -- +(2,2) --+(4,0);

\draw [dashed] (4,0) node[anchor=north] {$y$} -- +(0,1) -- +(-4, 1) node[anchor=east] {$S_t(x,y)$}; 

\draw [decoration={brace,mirror,raise=0.5cm},decorate] (1,0) -- +(2,0) node[pos=0.5,anchor=north,yshift=-0.55cm] {$p$}; 
\end{tikzpicture}
\caption{Fuzzy $\tnorm$-similarity relation $S_t$.}
\label{id:example:1}
\end{figure}
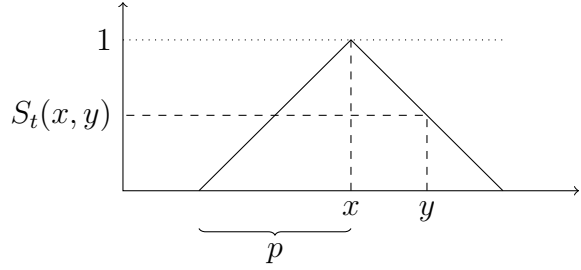

\begin{example}
Let's define a fuzzy $\tnorm$-similarity relation $S_t$
\begin{align}
S_t(x, y) = \max \left\{ 1 - \frac{|x - y|}{p}, 0 \right\}, \label{eq:S:max}
\end{align}
where $p > 0$ 
with the Łukasiewicz t-norm $\tnorm$
\begin{align}
	T(x,y) = \max \{0, x + y -1\}.
 \label{eq:tnorma:Luk}
\end{align}
Fig. \ref{id:example:1} presents a graphical representation of the \efn\ constructed with $S_t$.
\end{example}

\begin{theorem}
The relation $S_t$ (Eq. \ref{eq:S:max}) is a fuzzy $\tnorm$-similarity relation for the Łu\-ka\-sie\-wicz t-norm (Eq. \ref{eq:tnorma:Luk}).
\end{theorem}

\begin{proof}
The properties: reflexivity (Eq. \ref{eq:S:reflexivity}) and symmetry (\ref{eq:S:symmetry}) are a direct conclusion from the definition of the $S_t$ relation. The triangle property (Eq. \ref{eq:S:trojkat}) can be proofed with simple mathematics.
\end{proof}

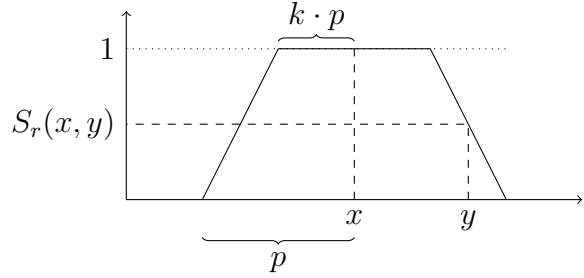
\begin{figure}
\centering
\begin{tikzpicture}
\draw[<->] (0,2.5) -- (0,0) -- (6,0);

\draw[dotted] (0,2) node[anchor=east] {$1$} -- +(5,0);

\draw[dashed] (3,0) node[anchor=north] {$x$} -- +(0,2);

\draw (1,0) -- +(1,2) -- +(3,2) --+(4,0);

\draw [dashed] (4.5,0) node[anchor=north] {$y$} -- +(0,1) -- +(-4.5, 1) node[anchor=east] {$S_r(x,y)$}; 

\draw [decoration={brace,mirror,raise=0.5cm},decorate] (1,0) -- +(2,0) node[pos=0.5,anchor=north,yshift=-0.55cm] {$p$}; 
\draw [decoration={brace,raise=0.1cm},decorate] (2,2) -- +(1,0) node[pos=0.5,anchor=south,yshift=0.1cm] {$k\cdot p$}; 
\end{tikzpicture}
\caption{Fuzzy $\tnorm$-similarity relation $S_r$.}
\label{id:example:r}
\end{figure}

\begin{example}
Let's define a fuzzy $\tnorm$-similarity relation $S_r$
\begin{align}
\ksR{S_r}{x}{y} = \min\left\{1, \max \left\{ 1 - \frac{|x - y|}{p}, 0 \right\} \cdot \frac{1}{1 - k}\right\}, \label{eq:S:trapezoidal}
\end{align}
where $p > 0$ and $0 \leqslant k < 1$
with the Łukasiewicz t-norm $\tnorm$
\begin{align}
	T(x,y) = \max \{0, x + y -1\}.
 \label{eq:tnorma:Luk}
\end{align}
Fig. \ref{id:example:r} presents a graphical representation of the \efn\ constructed with $S_r$.
\end{example}

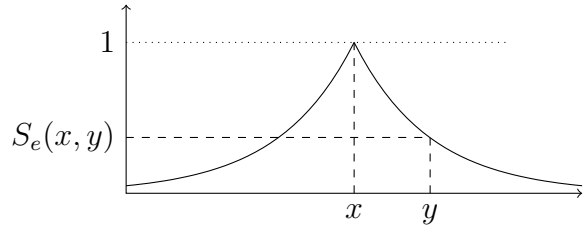
\begin{figure}
\centering
\begin{tikzpicture}
\draw[<->] (0,2.5) -- (0,0) -- (6,0);

\draw[dotted] (0,2) node[anchor=east] {$1$} -- +(5,0);

\draw[dashed] (3,0) node[anchor=north] {$x$} -- +(0,2);

\draw[smooth] (3.0,	2.000) -- (2.9,	1.810) -- (2.8,	1.637) -- (2.7,	1.482) -- (2.6,	1.341) -- (2.5,	1.213) -- (2.4,	1.098) -- (2.3,	0.993) -- (2.2,	0.899) -- (2.1,	0.813) -- (2.0,	0.736) -- (1.9,	0.666) -- (1.8,	0.602) -- (1.7,	0.545) -- (1.6,	0.493) -- (1.5,	0.446) -- (1.4,	0.404) -- (1.3,	0.365) -- (1.2,	0.331) -- (1.1,	0.299) -- (1.0,	0.271) -- (0.9,	0.245) -- (0.8,	0.222) -- (0.7,	0.201) -- (0.6,	0.181) -- (0.5,	0.164) -- (0.4,	0.149) -- (0.3,	0.134) -- (0.2,	0.122) -- (0.1,	0.110) -- (0.0,	0.100) ;
\draw[smooth] (3.0,	2.000) -- (3.1,	1.810) -- (3.2,	1.637) -- (3.3,	1.482) -- (3.4,	1.341) -- (3.5,	1.213) -- (3.6,	1.098) -- (3.7,	0.993) -- (3.8,	0.899) -- (3.9,	0.813) -- (4.0,	0.736) -- (4.1,	0.666) -- (4.2,	0.602) -- (4.3,	0.545) -- (4.4,	0.493) -- (4.5,	0.446) -- (4.6,	0.404) -- (4.7,	0.365) -- (4.8,	0.331) -- (4.9,	0.299) -- (5.0,	0.271) -- (5.1,	0.245) -- (5.2,	0.222) -- (5.3,	0.201) -- (5.4,	0.181) -- (5.5,	0.164) -- (5.6,	0.149) -- (5.7,	0.134) -- (5.8,	0.122) -- (5.9,	0.110) -- (6.0,	0.100) ;

\draw [dashed] (4,0) node[anchor=north] {$y$} -- +(0,0.74) -- +(-4, 0.74) node[anchor=east] {$S_e(x,y)$}; 

\end{tikzpicture}
\caption{Fuzzy $\tnorm$-similarity relation $S_e$ for $p=1$.}
\label{id:example:2}
\end{figure}

\begin{example}
Let's define a fuzzy $\tnorm$-similarity relation $S_e$ as:
\begin{align}
\ksR{S_e}{x}{y} = \exp \left( -\frac{|x - y|}{p} \right)
\label{eq:S2}
\end{align}
where $p > 1$ for the product t-norm $\tnorm$.
Fig. \ref{id:example:2} presents a graphical representation of the \efn\ constructed with $S_e$.
\end{example}

\begin{theorem}
The relation $S_e$ (Eq. \ref{eq:S2}) is a fuzzy $\tnorm$-similarity relation for the product t-norm $\tnorm$.
\end{theorem}

\begin{proof}
The properties: reflexivity (Eq. \ref{eq:S:reflexivity}) and symmetry (\ref{eq:S:symmetry}) are a direct conclusion from the definition of the $S_e$ relation. The triangle property (Eq. \ref{eq:S:trojkat}) can be proofed with simple mathematics.
\begin{align}
\ksR{S_e}{x}{y} \tnorm \ksR{S_e}{y}{z} & \leqslant \ksR{S_eg}{x}{z}\\
e^{ -\frac{|x - y|}{p}} \times e^{ -\frac{|y-z|}{p}} &\leqslant e^{ -\frac{|x - z|}{p}} \\
|x - y| +|y-z| &\geqslant |x - z| 
\end{align}
\end{proof}

\begin{figure}
\centering
\begin{tikzpicture}
\draw[<->] (0,2.5) -- (0,0) -- (6,0);

\draw[dotted] (0,2) node[anchor=east] {$1$} -- +(5,0);

\draw[dashed] (3,0) node[anchor=north] {$x$} -- +(0,2);

\draw[smooth] (3.0, 2.0) -- (3.1,	1.990) -- (3.2,	1.960) -- (3.3,	1.912) -- (3.4,	1.846) -- (3.5,	1.765) -- (3.6,	1.671) -- (3.7,	1.565) -- (3.8,	1.452) -- (3.9,	1.334) -- (4.0,	1.213) -- (4.1,	1.092) -- (4.2,	0.974) -- (4.3,	0.859) -- (4.4,	0.751) -- (4.5,	0.649) -- (4.6,	0.556) -- (4.7,	0.471) -- (4.8,	0.396) -- (4.9,	0.329) -- (5.0,	0.271) -- (5.1,	0.221) -- (5.2,	0.178) -- (5.3,	0.142) -- (5.4,	0.112) -- (5.5,	0.088) -- (5.6,	0.068) -- (5.7,	0.052) -- (5.8,	0.040) -- (5.9,	0.030) -- (6.0,	0.022);

\draw[smooth] (3.0, 2.0) -- (2.9,	1.990) -- (2.8,	1.960) -- (2.7,	1.912) -- (2.6,	1.846) -- (2.5,	1.765) -- (2.4,	1.671) -- (2.3,	1.565) -- (2.2,	1.452) -- (2.1,	1.334) -- (2.0,	1.213) -- (1.9,	1.092) -- (1.8,	0.974) -- (1.7,	0.859) -- (1.6,	0.751) -- (1.5,	0.649) -- (1.4,	0.556) -- (1.3,	0.471) -- (1.2,	0.396) -- (1.1,	0.329) -- (1.0,	0.271) -- (0.9,	0.221) -- (0.8,	0.178) -- (0.7,	0.142) -- (0.6,	0.112) -- (0.5,	0.088) -- (0.4,	0.068) -- (0.3,	0.052) -- (0.2,	0.040) -- (0.1,	0.030) -- (0.0,	0.022);

\draw [dashed] (4,0) node[anchor=north] {$y$} -- +(0,1.213) -- +(-4, 1.213) node[anchor=east] {$S_g(x,y)$}; 

\end{tikzpicture}
\caption{Fuzzy $\tnorm$-similarity relation $S_g$ for $p=1$.}
\label{id:example:S3}
\end{figure}
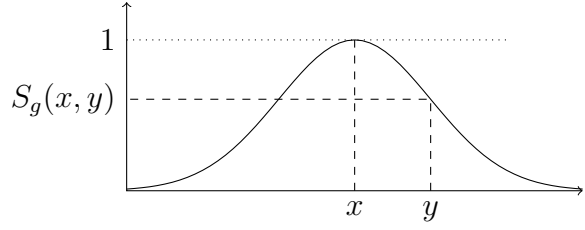

\begin{example}
Let's define a fuzzy $\tnorm$-similarity relation $S_g$ as:
\begin{align}
\ksR{S_g}{x}{y} = \exp \left( -\frac{(x - y)^2}{2p^2}\right)
\label{eq:S3}
\end{align}
where $p > 0$ for the product t-norm.
Fig. \ref{id:example:S3} presents a graphical representation of the \efn\ constructed with $S_g$.
\end{example}
\begin{theorem}
The relation $S_g$ (Eq. \ref{eq:S3}) is a fuzzy $\tnorm$-similarity relation for the product t-norm.
\end{theorem}

\begin{proof}
The properties: reflexivity (Eq. \ref{eq:S:reflexivity}) and symmetry (\ref{eq:S:symmetry}) are a direct conclusion from the definition of the $S_g$ relation. The triangle property (Eq. \ref{eq:S:trojkat}) can be proofed with simple mathematics.
\begin{align}
\ksR{S_g}{x}{y} \tnorm \ksR{S_g}{y}{z} & \leqslant \ksR{S_g}{x}{z}
\\
\exp \left( -\frac{(x - y)^2}{2p^2}\right)\times \exp \left( -\frac{(y - z)^2}{2p^2}\right) &\leqslant \exp \left( -\frac{(x - z)^2}{2p^2}\right) 
\\
(x - y)^2+(y-z)^2 &\geqslant (x-z)^2
\end{align}
\end{proof}

\begin{remark}
\Efn s constructed with relation $S_g$ and $S_e$ do not satisfy the classical definition of fuzzy numbers because their supports are infinite. 
“However, we do not see any theoretical nor conceptual reason for
that exclusion and we are convinced that this restriction is only of a technical
nature and stems from the $\alpha$-cut based arithmetics.” \cite{id:Holcapek2012Arithmethics-I}.
\end{remark}

For the application example (Sec. \ref{id:sec:fw}) we also define a 
fuzzy $\tnorm$-quasi-similarity relation.
\begin{definition}
\label{def:rel:tnorm-quasi-similarity}
A fuzzy $\tnorm$-quasi-similarity relation defined for a left continuous t-norm $\tnorm$ satisfies the following properties for all $x, y, z \in \mathbb{R}$:
\begin{align}
\ksR{S}{x}{x} & = 1, \label{eq:quasiS:reflexivity}\\
\ksR{S}{x}{y} \tnorm \ksR{S}{y}{z} &\leqslant \ksR{S}{x}{z}. \label{eq:S:quasiS:trojkat}
\end{align}
\end{definition}
The difference between a fuzzy $\tnorm$-quasi-similarity and a fuzzy $\tnorm$-similarity is that the former is not symmetric. 

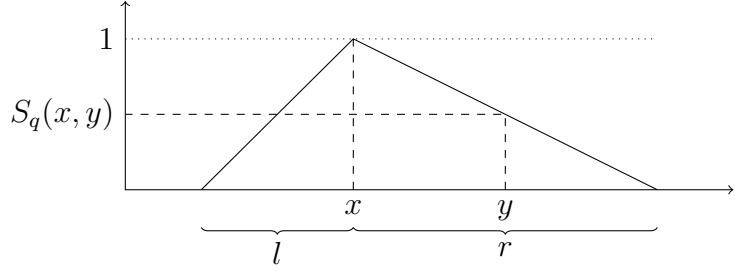
\begin{figure}
\centering
\begin{tikzpicture}
\draw[<->] (0,2.5) -- (0,0) -- (8,0);

\draw[dotted] (0,2) node[anchor=east] {$1$} -- +(7,0);

\draw[dashed] (3,0) node[anchor=north] {$x$} -- +(0,2);
\draw (1,0) -- +(2,2) --+(6,0);

\draw [dashed] (5,0) node[anchor=north] {$y$} -- +(0,1) -- +(-5, 1) node[anchor=east] {$S_q(x,y)$}; 

\draw [decoration={brace,mirror,raise=0.5cm},decorate] (1,0) -- +(2,0) node[pos=0.5,anchor=north,yshift=-0.55cm] {$l$}; 

\draw [decoration={brace,mirror,raise=0.5cm},decorate] (3,0) -- +(4,0) node[pos=0.5,anchor=north,yshift=-0.55cm] {$r$}; 

\end{tikzpicture}
\caption{Fuzzy $\tnorm$-quasi-similarity relation $S_q$.}
\label{id:example:S_q}
\end{figure}

\begin{example}
Let's define a fuzzy $\tnorm$-quasi-similarity relation $S_q$
\begin{align}
S_q(x, y) = 
\begin{cases}
\max \left\{ 1 - \frac{|x - y|}{l}, 0 \right\}, & x > y \\
\max \left\{ 1 - \frac{|x - y|}{r}, 0 \right\}, & x \leqslant y
\end{cases}
\label{eq:quasiS:max}
\end{align}
where $l > 0$ and $r > 0$ 
with the Łukasiewicz t-norm $\tnorm$.
Fig. \ref{id:example:S_q} presents a graphical representation of the \efn\ constructed with $S_q$.
\end{example}

\subsection{Arithmetics with \efn s}
The paper \cite{id:Holcapek2012Arithmethics-I} proves arithmetical operations on \efn s.
For $S \in \left\{S_t, S_r, S_e, S_g\right\}$: 
\begin{align}
x_{S\left(p_x\right)} + y_{S\left(p_y\right)} & = (x + y)_{S(\max\left\{p_x, p_y\right\})}, \label{eq:S:additioon} \\
x_{S\left(p_x\right)} \cdot y_{S\left(p_y\right)} & = (x \cdot y)_{S(\max\left\{p_x, p_y\right\})},\label{eq:S:multiplication}\\
- \left(x_S\right) & = \left(-x\right)_S, \\
\left(x_S\right)^{-1} & = \left(1/x\right)_S, \text{ for } x_S \neq 0_S. 
\end{align}

In \eqref{eq:S:additioon} and \eqref{eq:S:multiplication}, the $\max$ operator is used. The relations $S_t$, $S_r$, $S_e$, and $S_g$ are defined in such a way that $\max$ can be used. In general, a closure of two relations is used \cite{id:Holcapek2012Arithmethics-I}.

A very interesting feature of \efn s needs highlighting. In \efn s, there is not a unique zero and one. There are multiple zeroes and ones. All zeroes are constructed as extension of $x = 0$ with different vagueness. The same applies to one. The set of \efn s with addition, multiplication, identity elements 1 and 0 is not a algebraic field, it is a MI-monoid (many identities monoid) \cite{id:Holcapek2012Arithmethics-I}. It may seem strange, but it is completely intuitive, because we humans understand a fuzzy zero, eg. „more or less zero”, „precisely zero”. It is also counterintuitive that arithmetic manipulation with vague values should have crisp 0 and crisp 1.
For details and mathematical background please refer to \cite{id:Holcapek2012Arithmethics-I}.

We use the same approach as \cite{id:Holcapek2012Arithmethics-I} to define addition and multiplication of \efn s constructed with a fuzzy $\tnorm$-quasi-similarity relation. Thus:
\begin{align}
x_{S_q\left(l_x, r_x\right)} + y_{S_q\left(l_y, r_y\right)} & = (x + y)_{S_q(\max\left\{l_x, l_y\right\}, \max\left\{r_x, r_y\right\})}, \\
x_{S_q\left(l_x, r_x\right)} \cdot y_{S_q\left(l_y, r_y\right)} & = (x \cdot y)_{S_q(\max\left\{l_x, l_y\right\}, \max\left\{r_x, r_y\right\})}.
\end{align}

\subsection{Comparison operators for \efn s}

The papers \cite{id:Stepnicka2020Properties} define the $\leqslant$ operator for \efn s to return either an \efn\ for zero $0_S$ or an \efn\ for one $1_S$. 
In this paper, we need an operator that returns a value from an interval $[0,1]$. 
In following section, we define $\eqS, \lS, \leqS, \gS, \geqS$ operators for \efn s.

\begin{definition}
   A fuzzy binary relation $R$ is any fuzzy set of a Cartesian product of two sets $A, B$ of fuzzy objects:
\begin{align}
\ksR{R}{A}{B} \subseteq \left\{ \Membership(a, b) / (a,b) : a \in A \land b \in B\right\}
\end{align}
\end{definition}

\begin{definition}\label{id:def:eqS}
The equality operator $\eqS : \mathbb{E} \times \mathbb{E} \to [0,1]$, where $\mathbb{E}$ is a set of \efn s, is defined with the fuzzy $\tnorm$-similarity relation $S$ is a value of the $S$ relation for the bases of the \efn s: 
\begin{align}
\left(x_{S\left(p_x\right)} \eqS y_{S\left(p_y\right)}\right) & = 
S_{(\max\left\{p_x, p_y\right\})} (x,y).
\end{align}
\end{definition}
Equality relation is a fuzzy relation and its value is a number from an interval $[0,1]$.
Def. \ref{id:def:eqS} makes the equality ($\eqS$) transitive, because the fuzzy $\tnorm$-similarity relation $S$ satisfies the condition \eqref{eq:S:trojkat}.

We use the $\eqS$ relation for \efn s to define $\lS$, $\gS$, $\leqS$, $\geqS$ relations.
\begin{definition}\label{id:def:lS}
   The operator $\lS$ for two \efn s is defined as
\begin{align}
x_{S\left(p_x\right)} \lS y_{S\left(p_y\right)} & =
\begin{cases}
1 - \left(x_{S\left(p_x\right)} \eqS y_{S\left(p_y\right)}\right), & x \leqslant y,\\
0, & x > y.
\end{cases}
\end{align}
\end{definition}

\begin{definition}\label{id:def:gS}
   The operator $\gS$ for two \efn s is defined as
\begin{align}
x_{S\left(p_x\right)} \gS y_{S\left(p_y\right)} & =
\begin{cases}
1 - \left(x_{S\left(p_x\right)} \eqS y_{S\left(p_y\right)}\right), & x \geqslant y, \\
0, & x < y.
\end{cases}
\end{align}
\end{definition}

\begin{definition}\label{id:def:leqS}
   The operator $\leqS$ for two \efn s is defined as
\begin{align}
x_{S\left(p_x\right)} \leqS y_{S\left(p_y\right)} & =
\begin{cases}
1, & x \leqslant y,\\
\left(x_{S\left(p_x\right)} \eqS y_{S\left(p_y\right)}\right), & x > y.
\end{cases}
\end{align}
\end{definition}

\begin{definition}\label{id:def:geqS}
   The operator $\geqS$ for two \efn s is defined as
\begin{align}
x_{S\left(p_x\right)} \geqS y_{S\left(p_y\right)} & =
\begin{cases}
1, & x \geqslant y,\\
\left(x_{S\left(p_x\right)} \eqS y_{S\left(p_y\right)}\right), & x < y.
\end{cases}
\end{align}
\end{definition}

\begin{example}\label{id:ex:S_t}
Let's define three \efn s with the $S_t$ and the Łukasiewicz t-norm for crisp values: 
$a = 0$ (for $p_a = 1$), 
$b = 2$ (for $p_b = 3$),
$c = 3$ (for $p_c = 0.5$). 
The graphical representation of the numbers is presented in Fig. \ref{id:fig:ex:S_t}.
\begin{enumerate}
\item
The value of $a_{S_t(p_a)} \eqS b_{S_t(p_b)}$ is $S_{t(\max\{p_a, p_b\})} (a,b) = 
\max \left\{ 1 - \frac{|a - b|}{\max\{p_a, p_b\}}, 0 \right\} = 
\frac{1}{3}$.
Respectively: 
$a_{S_t(p_a)} \lS b_{S_t(p_b)} = \frac{2}{3}$,
$a_{S_t(p_a)} \leqS b_{S_t(p_b)} = 1$,
$a_{S_t(p_a)} \gS b_{S_t(p_b)} = 0$,
$a_{S_t(p_a)} \geqS b_{S_t(p_b)} = \frac{1}{3}$.

\item 
The value of  $b_{S_t(p_b)} \eqS c_{S_t(p_c)}$ is $S_{t(\max\{p_b, p_c\})} (b,c) 
= \max \left\{ 1 - \frac{|b - c|}{\max\{p_b, p_c\}}, 0 \right\} 
= \frac{2}{3}$.
Respectively: 
$b_{S_t(p_b)} \lS c_{S_t(p_c)} = \frac{1}{3}$,
$b_{S_t(p_b)} \leqS c_{S_t(p_c)} = 1$,
$b_{S_t(p_b)} \gS c_{S_t(p_c)} = 0$,
$b_{S_t(p_b)} \geqS c_{S_t(p_c)} = \frac{2}{3}$.

\item 
The value of $a_{S_t(p_a)} \eqS c_{S_t(p_c)}$ is $S_{t(\max\{p_a, p_c\})} (a,c) 
= \max \left\{ 1 - \frac{|a - c|}{\max\{p_a, p_c\}}, 0 \right\} 
= 0$.
Respectively: 
$a_{S_t(p_a)} \lS   c_{S_t(p_c)} = 1$,
$a_{S_t(p_a)} \leqS c_{S_t(p_c)} = 1$,
$a_{S_t(p_a)} \gS   c_{S_t(p_c)} = 0$,
$a_{S_t(p_a)} \geqS c_{S_t(p_c)} = 0$.
\end{enumerate}
\end{example}

\begin{figure}
\centering
\begin{tikzpicture}
\begin{axis}[legend style={at={(1.1,1)},anchor=north west}]
\addplot [domain=-2:6, samples=100] {  max(1 - abs(0 - x)/1, 0) }; \addlegendentry{$a_S = 0_{S_t(1)}$};
\addplot [domain=-2:6, samples=100,dashed] {  max(1 - abs(2 - x)/3, 0) }; \addlegendentry{$b_S = 2_{S_t(3)}$};
\addplot [domain=-2:6, samples=200,dashdotted] {  max(1 - abs(3 - x)/0.5, 0) }; \addlegendentry{$c_S = 3_{S_t(0.5)}$};
\end{axis}
\end{tikzpicture}
\caption{Three \efn s defined with the $S_t$ $\tnorm$-similarity relation and the Łukasiewicz t-norm.}
\label{id:fig:ex:S_t}
\end{figure}

\begin{figure}
\centering
\begin{tikzpicture}
\begin{axis}[legend style={at={(1.1,1)},anchor=north west}]
\addplot [domain=-5:10, samples=200] { min (1, 2.0 * max(1 - abs(0 - x)/3, 0) ) }; \addlegendentry{$a_S = 0_{S_r(3)}$};
\addplot [domain=-5:10, samples=100,dashed] { min(1, 2.0 * max(1 - abs(3 - x)/5, 0) ) }; \addlegendentry{$b_S = 3_{S_r(5)}$};
\addplot [domain=-5:10, samples=100,dashdotted] {min(1, 2.0 *  max(1 - abs(7 - x)/0.5, 0) ) }; \addlegendentry{$c_S = 7_{S_r(0.5)}$};
\end{axis}
\end{tikzpicture}
\caption{Three \efn s defined with the $S_r$ fuzzy $\tnorm$-similarity relation and the Łukasiewicz t-norm.}
\label{id:fig:ex:Sr}
\end{figure}
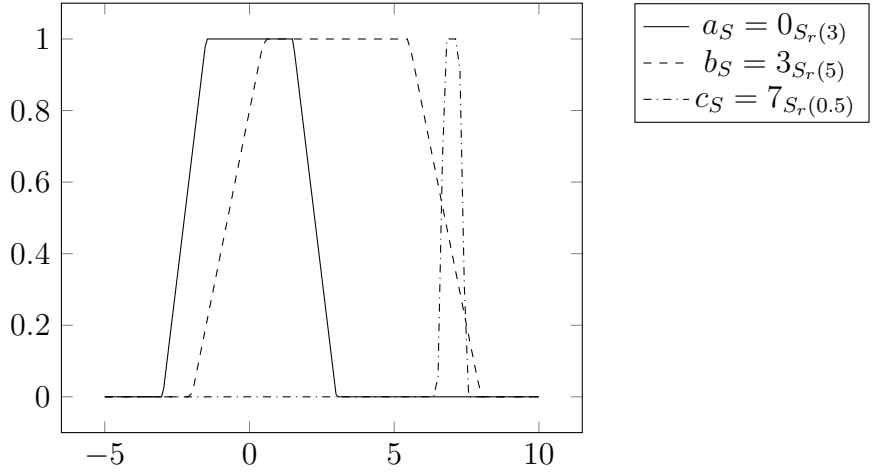

\begin{example}\label{id:ex:S1}
Let's define three \efn s with the $S_r$ and the Łukasiewicz t-norm for crisp values:
$a = 0$ (for $p_a = 3$), 
$b = 3$ (for $p_b = 5$),
$c = 7$ (for $p_c = 0.5$),
$k = 0.5$.
The numbers are presented in Fig. \ref{id:fig:ex:Sr} 

\begin{enumerate}
\item 
The value of $a_{S_r(p_a)} \eqS \allowdisplaybreaks b_{S_r(p_b)}$ is by definition $S_{r(\max\{p_a, p_b\})} (a,b) 
= \allowdisplaybreaks \min\left\{1,\max \left\{ 1 - \frac{|a - b|}{\max\{p_a, p_b\}}, 0 \right\} \cdot \frac{1}{1-k}\right\} 
= \frac{4}{5}$.
Respectively: 
$a_{S_t(p_a)} \lS   b_{S_t(p_b)} = \frac{1}{5}$,
$a_{S_t(p_a)} \leqS b_{S_t(p_b)} = 1$,
$a_{S_t(p_a)} \gS   b_{S_t(p_b)} = 0$,
$a_{S_t(p_a)} \geqS b_{S_t(p_b)} = \frac{4}{5}$.

\item 
The value of $b_{S_r(p_b)} \eqS c_{S_r(p_c)}$ is by definition $S_{r(\max\{p_b, p_c\})} (b,c) 
= \min\left\{1,\max \left\{ 1 - \frac{|b - c|}{\max\{p_b, p_c\}}, 0 \right\}  \cdot \frac{1}{1-k}\right\} 
= \frac{2}{5}$.
Respectively: 
$b_{S_t(p_b)} \lS   c_{S_t(p_c)} = \frac{3}{5}$,
$b_{S_t(p_b)} \leqS c_{S_t(p_c)} = 1$,
$b_{S_t(p_b)} \gS   c_{S_t(p_c)} = 0$,
$b_{S_t(p_b)} \geqS c_{S_t(p_c)} = \frac{2}{5}$.

\item 
The value of $a_{S_t(p_a)} \eqS c_{S_t(p_b)}$ is by definition $S_{r(\max\{p_a, p_c\})} (a,c) 
= \min\left\{1,\max \left\{ 1 - \frac{|a - c|}{\max\{p_a, p_c\}}, 0 \right\}   \cdot \frac{1}{1-k}\right\} 
= 0$.
Respectively:
$a_{S_t(p_a)} \lS   c_{S_t(p_c)} = 1$,
$a_{S_t(p_a)} \leqS c_{S_t(p_c)} = 1$,
$a_{S_t(p_a)} \gS   c_{S_t(p_c)} = 0$,
$a_{S_t(p_a)} \geqS c_{S_t(p_c)} = 0$.
\end{enumerate}
\end{example}

\begin{figure}
\centering
\begin{tikzpicture}
\begin{axis}[legend style={at={(1.1,1)},anchor=north west}]
\addplot [domain=-5:10, samples=100] { exp(-abs(0 - x)/1) }; \addlegendentry{$a_S = 0_{S_e(1)}$};
\addplot [domain=-5:10, samples=100,dashed] {  exp(-abs(2 - x)/3) }; \addlegendentry{$b_S = 2_{S_e(3)}$};
\addplot [domain=-5:10, samples=100,dashdotted] {  exp(-abs(5 - x)/0.5)}; \addlegendentry{$c_S = 5_{S_e(0.5)}$};
\end{axis}
\end{tikzpicture}
\caption{Three \efn s defined with the $S_e$ fuzzy $\tnorm$-similarity relation and the Łukasiewicz t-norm.}
\label{id:fig:ex:S_e}
\end{figure}
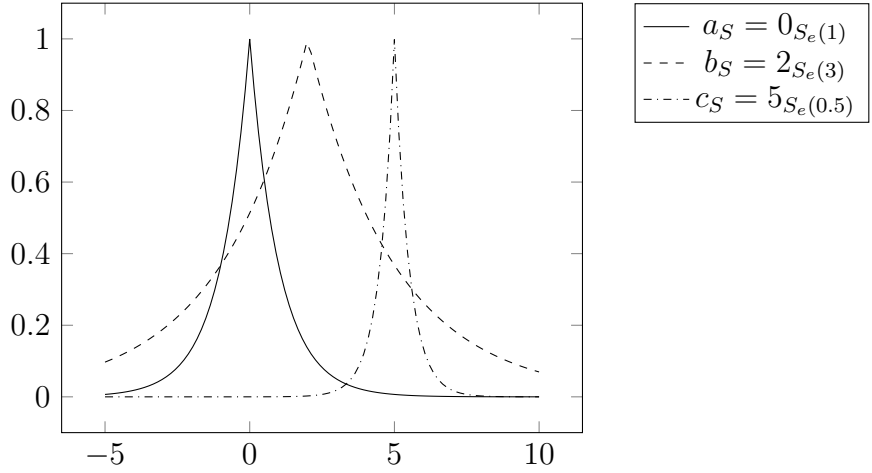

\begin{example}\label{id:ex:S_e}
Let's define three \efn s with the $S_e$ and the product t-norm for crisp values:
$a = 0$ (for $p_a = 1$), 
$b = 2$ (for $p_b=3$),
$c = 5$ (for $p_c=0.5$). 
The numbers are presented in Fig. \ref{id:fig:ex:S_e}.
\begin{enumerate}
\item 
The value of  $a_{S_e(p_a)} \eqS b_{S_e(p_b)}$ is $S_{e(\max\{p_a, p_b\})} (a,b) = \exp \left( -\frac{|a - b|}{\max\{p_a, p_b\}} \right) 
\approx 0.513 $.
Respectively:
$a_{S_e(p_a)} \lS   b_{S_e(p_b)} \approx 1 - 0.513 = 0.487$,
$a_{S_e(p_a)} \leqS b_{S_e(p_b)} = 1$,
$a_{S_e(p_a)} \gS   b_{S_e(p_b)} = 0$,
$a_{S_e(p_a)} \geqS b_{S_e(p_b)} \approx 0.513$.
\item 
The value of  $a_{S_e(p_a)} \eqS c_{S_e(p_c)}$ is $S_{e(\max\{p_a, p_c\})} (a,c) = \exp \left( -\frac{|a - c|}{\max\{p_a, p_c\}} \right) 
\approx 0.0067 $.
Respectively: 
$a_{S_e(p_a)} \lS   c_{S_e(p_c)} \approx 1 - 0.0067 = 0.9933$,
$a_{S_e(p_a)} \leqS c_{S_e(p_c)} = 1$,
$a_{S_e(p_a)} \gS   c_{S_e(p_c)} = 0$,
$a_{S_e(p_a)} \geqS c_{S_e(p_c)} \approx 0.0067$.

\item 
The value of $b_{S_e(p_b)} \eqS c_{S_e(p_c)}$ is $S_{e(\max\{p_b, p_c\})} (b,c) = \exp \left( -\frac{|b - c|}{\max\{p_b, p_c\}} \right) 
\approx 0.368  $.
Respectively: 
$b_{S_e(p_b)} \lS   c_{S_e(p_c)} \approx 1 - 0.368 = 0.632$,
$b_{S_e(p_b)} \leqS c_{S_e(p_c)} = 1$,
$b_{S_e(p_b)} \gS   c_{S_e(p_c)} = 0$,
$b_{S_e(p_b)} \geqS c_{S_e(p_c)} \approx 0.368$.
\end{enumerate}
\end{example}

\begin{figure}
\centering
\begin{tikzpicture}
\begin{axis}[legend style={at={(1.1,1)},anchor=north west}]
\addplot [domain=-5:10, samples=100] { exp (- (x - 0)^2 / (2 * 1^2)) }; \addlegendentry{$a_S = 0_{S_g(1)}$};
\addplot [domain=-5:10, samples=100,dashed] { exp (- (x - 2)^2 / (2 * 3^2)) }; \addlegendentry{$b_S = 2_{S_g(3)}$};
\addplot [domain=-5:10, samples=100,dashdotted] { exp (- (x - 5)^2 / (2 * 0.5^2)) }; \addlegendentry{$c_S = 5_{S_g(0.5)}$};
\end{axis}
\end{tikzpicture}
\caption{Three \efn s defined with the $S_g$ fuzzy $\tnorm$-similarity relation and the Łukasiewicz t-norm.}
\label{id:fig:ex:S3}
\end{figure}
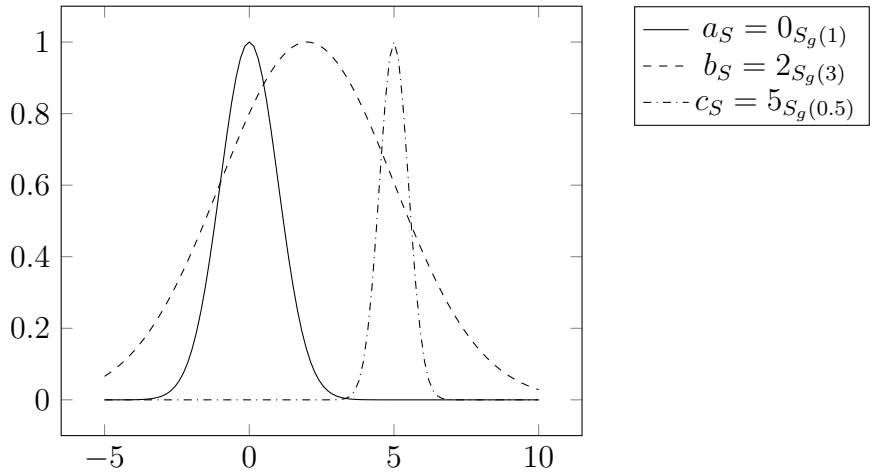

\begin{example}\label{id:ex:S3}
Let's define three \efn s with the $S_g$ and the product t-norm for crisp values:
$a = 0$ (for $p_a = 1$), 
$b = 2$ (for $p_b=2$),
$c = 5$ (for $p_c=0.5$). 
The numbers are presented in Fig. \ref{id:fig:ex:S3}. 
\begin{enumerate}
\item 
The value of $a_{S_g(p_a)} \eqS b_{S_g(p_b)}$ is $S_{g(\max\{p_a, p_b\})} (a,b) = \exp \left(-\frac{(a - b)^2}{2(\max\{p_a, p_b\})^2} \right) 
\approx 0.8007$. 
Respectively:
$a_{S_g(p_a)} \lS   b_{S_g(p_b)} \approx 1 - 0.8007 = 0.1993$,
$a_{S_g(p_a)} \leqS b_{S_g(p_b)} = 1$,
$a_{S_g(p_a)} \gS   b_{S_g(p_b)} = 0$,
$a_{S_g(p_a)} \geqS b_{S_g(p_b)} \approx 0.8007$,

\item 
The value of $a_{S_g(p_a)} \eqS c_{S_g(p_c)}$ is $S_{g(\max\{p_a, p_c\})} (a,c) = \exp \left(-\frac{(a - c)^2}{2(\max\{p_a, p_c\})^2} \right) 
\approx 0.000003727$. 
Respectively:
$a_{S_g(p_a)} \lS   c_{S_g(p_c)} \approx 1 - 0.000003727 = 0.999996273$,
$a_{S_g(p_a)} \leqS c_{S_g(p_c)} = 1$,
$a_{S_g(p_a)} \gS   c_{S_g(p_c)} = 0$,
$a_{S_g(p_a)} \geqS c_{S_g(p_c)} \approx 0.000003727$.

\item 
The value of $b_{S_g(p_b)} \eqS c_{S_g(p_c)}$ is $S_{g(\max\{p_b, p_c\})} (b,c) = \exp \left(-\frac{(b - c)^2}{2(\max\{p_b, p_c\})^2} \right) 
\approx 0.6065$. 
Respectively:
$b_{S_g(p_b)} \lS   c_{S_g(p_c)} \approx 1 - 0.6065 = 0.3935$,
$b_{S_g(p_b)} \leqS c_{S_g(p_c)} = 1$,
$b_{S_g(p_b)} \gS   c_{S_g(p_c)} = 0$,
$b_{S_g(p_b)} \geqS c_{S_g(p_c)} \approx 0.6065$.

\end{enumerate}
\end{example}

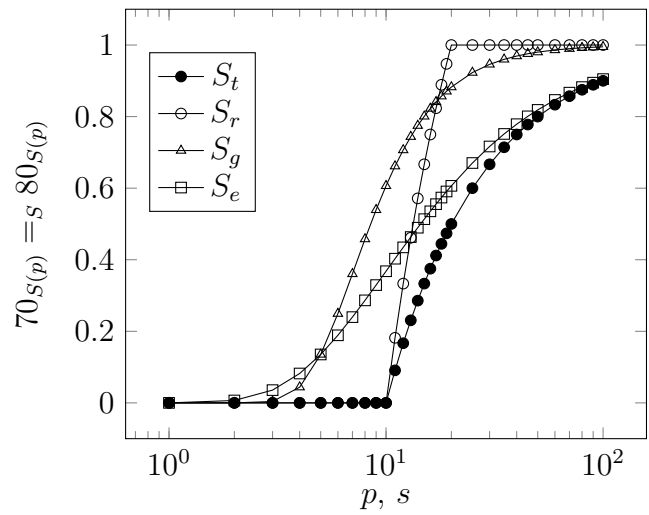
\begin{figure}
\centering
\begin{tikzpicture}
\begin{semilogxaxis}[
		xlabel={$p$, $s$},
		ylabel={$70_{S(p)} \eqS 80_{S(p)}$},
		legend style={at={(0.25,0.9)}}
    ]
    \addplot[mark=*] table[x=rozmycie,y=triangular]  {data.data}; \addlegendentry{$S_t$};
    \addplot[mark=o] table[x=rozmycie,y=trapezoidal]  {data.data}; \addlegendentry{$S_r$};
    \addplot[mark=triangle] table[x=rozmycie,y=gaussian]  {data.data}; \addlegendentry{$S_g$};
    \addplot[mark=square] table[x=rozmycie,y=expabs]  {data.data}; \addlegendentry{$S_e$};
\end{semilogxaxis}
\end{tikzpicture}
\caption{Comparison of values of the relation $70_{S(p)} \eqS 80_{S(p)}$ for various values of the $p$ parameter and $S \in \left\{S_t, S_r, S_g, S_e\right\}$.}
\label{fig:comparison}
\end{figure}

\begin{example}
Fig. \ref{fig:comparison} presents the comparison of impact of the fuzzification parameter $p$ on the value of the $\eqS$ operator.
The cores of the \efn s are constants 70 and 80. The higher the value of $p$ (fuzzification), the equality of numbers is higher. 
This is fully with concordance with the intuition that if two non identical fuzzy numbers are very fuzzy, they can be treated as almost equal and can be used to present the same fuzzy term.
The equality operator $\eqS$ can return 1 for two fuzzy numbers with different cores only for the trapezoidal $S_r$ relation of four presented in this paper. 
\end{example}

\section{Application examples}
The implementation of \efn s and the source code for the examples below are available from the free public GitHub repository: \url{github.com/ksiminski/ksi-extensional-fuzzy-numbers}.

\subsection{Sorting}
Fig. \ref{id:pc:insertion_sort} presents a pseudocode for the insertion sort algorithm for \efn s. The $\lS$ operator returns a value from the interval $[0,1]$. 
In order to sort fuzzy numbers a threshold $\xi$ is needed because a decision has to be made whether to swap two \efn s. The threshold is a parameter of the sorting algorithm.

\begin{figure}
\centering
\begin{lstlisting}
procedure insertion_sort (A, @$\xi$@)
@\emph{// }@A[1..n] @\emph{ : array of \efn s}@
@\emph{// }$\xi$\emph{ : threshold for the $\lS$ operator for \efn s}@

   for i @$\leftarrow$@ 2 to n do
      minimum @$\leftarrow$@ A[i];
      j @$\leftarrow$@ i - 1;
      while j > 0 and (minimum @$\lS$@ A[j]) > @$\xi$@ do @\label{id:pc:insertion_sort:less}@
         A[j + 1] @$\leftarrow$@ A[j];
         j @$\leftarrow$@ j - 1;
      end while;
      A[j + 1] @$\leftarrow$@ minimum;
   end for;
end procedure;
\end{lstlisting}
\caption{Pseudocode of the insertion sort algorithm for \efn s.}
\label{id:pc:insertion_sort}
\end{figure}

The impact of the $\xi$ threshold parameter can be seen in the following example.

\begin{figure}
\centering
\begin{tikzpicture}
\begin{axis}[width=15cm,height=5cm,legend style={at={(1.1,1)},anchor=north west}]

\foreach \m/\s in {5.5/0.5, 6/3, 4.5/1, -4.3/1, 5.6/1, 0.4/2, 7.7/1, 4.2/1.5} 
{
  \addplot [domain=-6:10, samples=200] { exp (- (x - \m)^2 / (2 * \s^2)) };    
}
\end{axis}

\end{tikzpicture}
\caption{Gaussian \efn s used for sorting.}
\end{figure}

\begin{figure}
\centering
\begin{tikzpicture}
    
\begin{axis}[legend style={at={(1,0)},anchor=south east}, minor x tick num = 1, minor y tick num = 4, xlabel={array cell index}, ylabel={fuzzy number core}]
  \addplot[mark=square] coordinates {(1, -4.3) (2, 5.5) (3, 6) (4, 4.5) (5, 0.4) (6, 5.6) (7, 4.2) (8, 7.7)}; \addlegendentry{$\xi = 0.9$};
  \addplot[mark=o] coordinates { (1, -4.3) (2, 0.4) (3, 5.5) (4, 6) (5, 4.5) (6, 4.2) (7, 5.6) (8, 7.7)}; \addlegendentry{$\xi = 0.5$};
  \addplot[mark=*] coordinates {(1,-4.3) (2,0.4) (3,4.2) (4,4.5) (5,5.5) (6,5.6) (7,6) (8,7.7)}; \addlegendentry{$\xi = 0.0$};
\end{axis}

\end{tikzpicture}
\caption{Sorting of Gaussian \efn s for various values of a threshold $\xi$. The plot presents the location of cores of Gaussian \efn s in cells of the sorted array. The plot does not present the vagueness of numbers.}
\label{fig:sorting}
\end{figure}
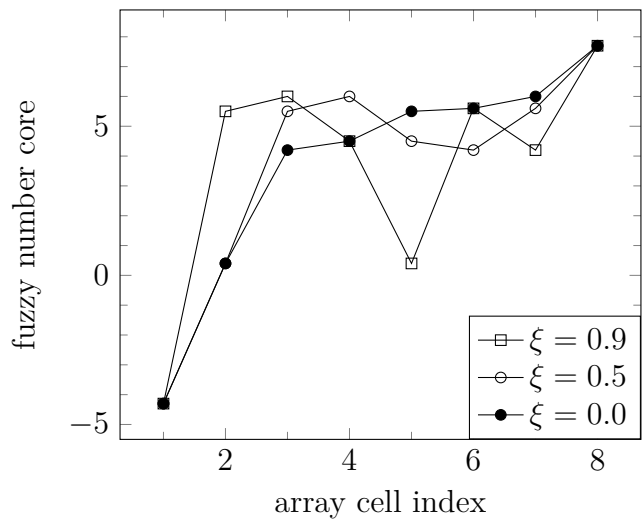

Initial array of Gaussian \efn s numbers:
$$ (5.5, 0.5) (6, 3) (4.5, 1) (-4.3, 1) (5.6, 1) (0.4, 2) (7.7, 1) (4.2, 1.5).$$
Results of sorting for various $\xi$'s:
\begin{itemize}
\item $\xi = 0.9$ : $(-4.3, 1) (5.5, 0.5) (6, 3) (4.5, 1) (0.4, 2) (5.6, 1) (4.2, 1.5) (7.7, 1) $
\item $\xi = 0.5$ : $ (-4.3, 1) (0.4, 2) (5.5, 0.5) (6, 3) (4.5, 1) (5.6, 1) (4.2, 1.5) (7.7, 1)  $
\item $\xi = 0.0$ : $ (-4.3, 1) (0.4, 2) (4.2, 1.5) (4.5, 1) (5.5, 0.5) (5.6, 1) (6, 3) (7.7, 1)  $
\end{itemize}
For $\xi=0$, the vagueness for fuzzy numbers has no influence on the result of sorting. For higher $\xi$'s, some numbers cannot be swapped in the sorting algorithm due to their significant vagueness.
The graphical representation of the orderings is presented in Fig. \ref{fig:sorting}.

\subsection{The shortest paths in a fuzzy weighted graph}
\label{id:sec:fw}

The problem stated in this section is a fuzzy version of the very well known problem in weighted graphs. It is commonly solved with the Dijkstra's algorithm (all paths from one node, no negative edges allowed), Bellman-Ford algorithm (all paths from one node,  negative edges allowed), Floyd-Warshall algorithm (all paths between any pair of vertices).

The problems in application of fuzzy numbers to the shortest path problem are \cite{id:Hernandes2007Shortest}:
\begin{itemize}
\item costs without existing paths \cite{id:Okada2000Shortest,id:Hernandes2007Shortest},
\item a fuzzy set of nondominated paths \cite{id:Okada2000Shortest} as a result,
\item limitation to graphs with non-negative edge weights.
\end{itemize}

The main problem in this task is the ordering of fuzzy numbers. This is solved with assignment an index value to fuzzy numbers. 
This is not trivial because commonly indices need elaboration of integrals (Yager's index \cite{id:Yager1978Ranking,id:Yager1980Choosing,id:Yager1981Procedure}, Liou-Wang index \cite{id:Liou1992Ranking}, García-Lamata index \cite{id:Garcia2005Fuzzy}). 
Some techniques avoid integrals \cite{id:Okada2000Shortest,id:Kuchta2002Generalisation}.
Dubois and Prade proposed possibility index \cite{id:Dubois1983Ranking} and Nayeem and Pal acceptability index 
\cite{id:Nayeem2005Shortest}.
Indices proposed by Yager, Liou and Wang, García and Lamata, Dubois, and Prade return a single solution.
Algorithms based on the 
Okada and Soper index and the Nayeem and Pal acceptability index return a set of solutions. 
Some authors apply for this problem genetic algorithms \cite{id:Lin2021Genetic,id:Hassanzadeh2013Genetic} or dynamic programming \cite{id:Mahdavi2009Dynamic,id:Tajdin2010Computing}.

In following sections, we present our approach in comparison with the results elaborated by the techniques mentioned above. We use quasi-similarity relation to enable asymmetric triangular fuzzy numbers because all techniques above apply asymmetric triangular fuzzy numbers to represent edge weights. 

We apply the Floyd-Warshall algorithm, whose pseudocode in presented in Fig. \ref{id:pc:fl}.

\begin{figure}
\begin{lstlisting}
procedure Floyd_Washall (@$G=(\mathbb{V}, \mathbb{E}), w, \xi$@)
   @\emph{// }$w$\emph{ – edge weights}@
   @\emph{// }$\xi$\emph{ – threshold for the comparison operator}@
   @\emph{// }@d@\emph{ – array of distances}@
   @\emph{// }@p@\emph{ – array of predecessors}@
   
   n @$\leftarrow$@ @$|\mathbb{V}|$@; @\emph{// number of edges}@@\label{id:alg:fw:initialisation}@
   
   for i @$\leftarrow$@ 1 to n do  @\emph{// initialisation}@
      for j @$\leftarrow$@ 1 to n do
         p[i,j] @$\leftarrow$@ j;
         if i = j then
            d[i,j] @$\leftarrow 0$@;
         else if @$\exists w_{ij}$@ then  
            d[i,j] @$\leftarrow w_{ij}$@;
         else
            d[i,j] @$\leftarrow \infty$@;            
            p[i,j] @$\leftarrow \varnothing$@;
         end if;
      end for;
   end for;  
   
   for k @$\leftarrow$@ 1 to n do     @\emph{// elaboration of distances}@
      for i @$\leftarrow$@ 1 to n do
         if d[i, k] @$\neq\infty$@ then
            for j @$\leftarrow$@ 1 to n do
               if (d[i,k] + d[k,j] @$\lS$@ d[i,j]) > @$\xi$@ then
                  d[i,j] @$\leftarrow$@ d[i,k] + d[k,j]; @\label{id:alg:fw:actualisation:d}@
                  p[i,j] @$\leftarrow$@ p[i,k]; @\label{id:alg:fw:actualisation:p}@
               end if;
            end for;
         end if;
      end for;
   end for;   
end procedure;
\end{lstlisting}
\caption{The Floyd-Warshall algorithm for the shortest  paths between all pairs of nodes in a graph with fuzzy weights.}
\label{id:pc:fl}
\end{figure}

In the following examples, edge weights are asymmetric triangular fuzzy number in form $(m, \alpha, \beta)$, where $m$ stands for the core, $(m - \alpha, m + \beta)$ is the support interval \cite{id:Hernandes2007Shortest}.

\subsubsection{Example 1}
\label{sec:ex:1}

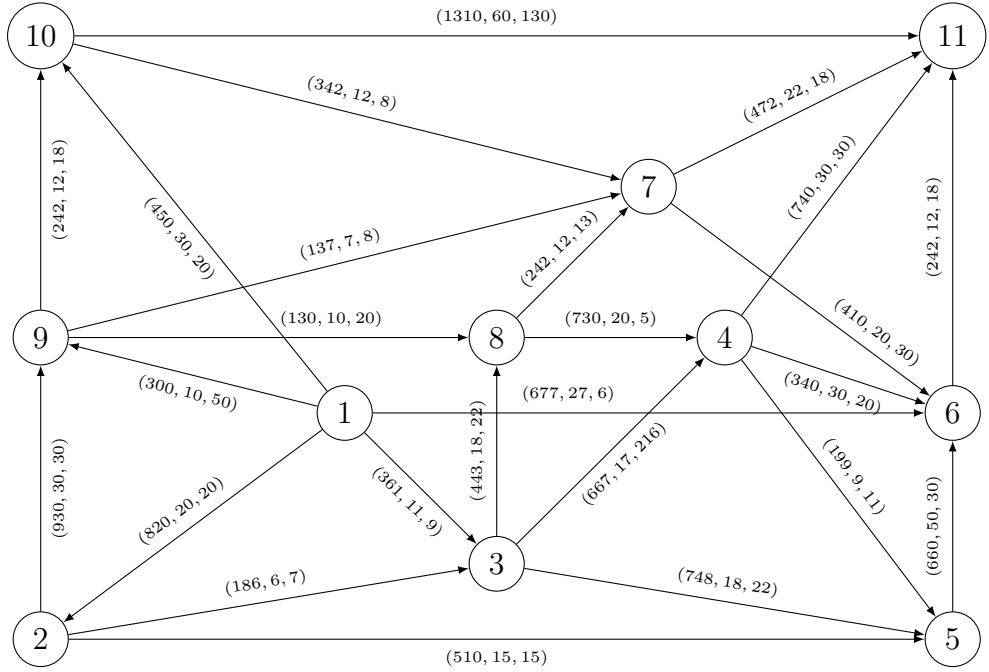
\begin{figure}
\centering 
\begin{tikzpicture}

\draw (0, 10) node (10) [circle,draw=black] {10};
\draw (4,  5) node  (1) [circle,draw=black]  {1};

\draw (0,  6) node  (9) [circle,draw=black]  {9};

\draw (8,  8) node  (7) [circle,draw=black]  {7};
\draw (6,  6) node  (8) [circle,draw=black]  {8};
\draw (6,  3) node  (3) [circle,draw=black]  {3};
\draw (0,  2) node  (2) [circle,draw=black]  {2};

\draw (9,  6) node  (4) [circle,draw=black]  {4};

\draw (12, 10) node (11) [circle,draw=black] {11};
\draw (12,  5) node  (6) [circle,draw=black]  {6};
\draw (12,  2) node  (5) [circle,draw=black]  {5};

\path[-latex] (1) edge node[above,sloped] {\tiny $(820 ,20 ,20  )$} (2);
\path[-latex] (1) edge node[below,sloped] {\tiny $(361 ,11 ,9   )$} (3);
\path[-latex] (1) edge node[near start,anchor=south west] {\tiny $(677 ,27 ,6   )$} (6);
\path[-latex] (1) edge node[below,sloped] {\tiny $(300 ,10 ,50  )$} (9);
\path[-latex] (1) edge node[below,sloped] {\tiny $(450 ,30 ,20  )$}(10);

\path[-latex] (2) edge node[above,sloped] {\tiny $(186 ,6  ,7   )$} (3);
\path[-latex] (2) edge node[below,sloped] {\tiny $(510 ,15 ,15  )$} (5);
\path[-latex] (2) edge node[below,sloped] {\tiny $(930 ,30 ,30  )$} (9);

\path[-latex] (3) edge node[below,sloped] {\tiny $(667 ,17 ,216 )$} (4);
\path[-latex] (3) edge node[above,sloped] {\tiny $(748 ,18 ,22  )$} (5);
\path[-latex] (3) edge node[above,sloped] {\tiny $(443 ,18 ,22  )$} (8);

\path[-latex] (4) edge node[above,sloped] {\tiny $(199 ,9  ,11  )$} (5);
\path[-latex] (4) edge node[below,sloped] {\tiny $(340 ,30 ,20  )$} (6);
\path[-latex] (4) edge node[above,sloped] {\tiny $(740 ,30 ,30  )$}(11);

\path[-latex] (5) edge node[above,sloped] {\tiny $(660 ,50 ,30  )$} (6);

\path[-latex] (6) edge node[above,sloped] {\tiny $(242 ,12 ,18  )$}(11);

\path[-latex] (7) edge node[above,sloped,near end] {\tiny $(410 ,20 ,30  )$} (6);
\path[-latex] (7) edge node[above,sloped] {\tiny $(472 ,22 ,18  )$}(11);

\path[-latex] (8) edge node[above,sloped] {\tiny $(730 ,20 ,5   )$} (4);
\path[-latex] (8) edge node[above,sloped] {\tiny $(242 ,12 ,13  )$} (7);

\path[-latex] (9) edge node[above,sloped] {\tiny $(137 ,7  ,8   )$} (7);
\path[-latex] (9) edge node[anchor=south west] {\tiny $(130 ,10 ,20  )$} (8);
\path[-latex] (9) edge node[below,sloped] {\tiny $(242 ,12 ,18  )$}(10);

\path[-latex] (10) edge node[above,sloped] {\tiny $(342 ,12 ,8   )$} (7);
\path[-latex] (10) edge node[above,sloped] {\tiny $(1310, 60, 130)$}(11);

\end{tikzpicture}
\caption{Graph with fuzzy weights from Example 1 \cite{id:Hernandes2007Shortest}.}
\label{fig:fuzzy-graph-1}
\end{figure}

This is an application of \efn s for the example 1 from \cite{id:Hernandes2007Shortest}. 
The graph is presented in Fig. \ref{fig:fuzzy-graph-1}. %
Node 1 is the starting node.

Tab. \ref{tab:example:1} is taken from \cite{id:Hernandes2007Shortest} with additional column (EFN) presenting lengths of paths elaborated with \efn s.
Bellow, we present non-direct paths elaborated for the threshold
$\xi = 0.3$:
\begin{itemize}
\item The path from node 1 to 4 has length $(1028, 17, 216)$. The path is $1 \xrightarrow{(361, 11, 9)} 3 \xrightarrow{(667, 17, 216)} 4$.
\item The path from node 1 to 11 has length $(902, 22, 50)$. The path is $1 \xrightarrow{(300, 10, 50)} 9 \xrightarrow{(130, 10, 20)} 7 \xrightarrow{(472, 22, 18)} 11$.
\end{itemize}

\begin{table}
\centering
\caption{The order relations for Example 1 are taken from \cite{id:Hernandes2007Shortest}.  Yager (Y),  Liou-Wang (LW),  García-Lamata (GL), Okada-Soper (OS),  Nayeem-Pal (NP),  Dubois-Prade (DP). The column EFN (\efn) presents path length elaborated with our approach.}
\label{tab:example:1}
\begin{tabular}{rlllp{3cm}}
	\toprule
	                                           end & shortest path            & \multicolumn{2}{c}{path cost} & order relation \\
	\cmidrule(lr){3-4} 
 node &                          & \cite{id:Hernandes2007Shortest} &  EFN                           & \\
	\midrule
  2 & $1 \to 2$                & $(820 ,20 ,20)  $ & $(820, 20, 20)  $  & all\\
  3 & $1 \to 3$                & $(361 ,11 ,9)   $ & $(361, 11, 9)   $  & all\\
  4 & $1 \to 3 \to 4  $        & $(1028, 28, 225)$ & $(1028, 17, 216)$ & all\\
  4 & $1 \to 9 \to 8 \to 4  $  & $(1167, 37, 63) $ & n/a                & OS $(\varepsilon=0)$ \\
  5 & $1 \to 3 \to 5   $       & $(1109, 29, 31) $ & $(1109, 18, 22) $  & all\\
  6 & $1 \to 6  $              & $(677 ,27 ,6)   $ & $(677, 27, 6)   $  & all\\
  7 & $1 \to 9 \to 7   $       & $(430 ,20 ,70)  $ & $(430, 10, 50)  $  & all\\
  8 & $1 \to 9 \to 8   $       & $(437 ,17 ,58)  $ & $(437, 10, 50)  $  &all\\
  9 & $1 \to 9  $              & $(300 ,10 ,50)  $ & $(300, 10, 50)  $  & all\\
 10 & $1 \to 10  $             & $(450 ,30 ,20)  $ & $(450, 30, 20)  $ & all\\
 11 & $1 \to 9 \to 7 \to 11  $ & $(902 ,42 ,88)  $ & $(902, 22, 50)  $  & Y, LW $(\lambda=1)$, GL $(\lambda=1,\delta=0)$, OS  $(\varepsilon=0, \varepsilon=0.5)$, EFN\\
 11 & $1 \to 6 \to 11  $       & $(919 ,39 ,24)  $ & n/a                & LW $(\lambda=0, \lambda=0.5)$, GL  $(\lambda\neq 1, \delta\neq 0)$, NP, DP\\
\bottomrule	       
\end{tabular}
\end{table}

\subsubsection{Example 2}
\label{sec:ex:2}

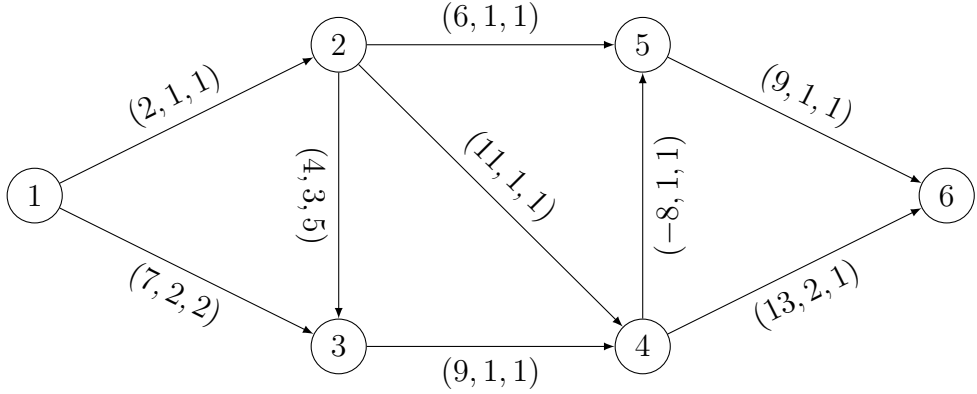
\begin{figure}
\centering 
\begin{tikzpicture}

\draw (4,  6) node  (1) [circle,draw=black]  {1};

\draw (8,  8) node  (2) [circle,draw=black]  {2};
\draw (8,  4) node  (3) [circle,draw=black]  {3};

\draw (12,  8) node  (5) [circle,draw=black]  {5};
\draw (12,  4) node  (4) [circle,draw=black]  {4};

\draw (16,  6) node  (6) [circle,draw=black]  {6};

\path[-latex] (1) edge node[above,sloped] {$(2, 1, 1)$} (2);
\path[-latex] (1) edge node[below,sloped] {$(7, 2, 2)$}   (3); 

\path[-latex] (2) edge node[below,sloped] {$(4, 3, 5)$}   (3);
\path[-latex] (2) edge node[above,sloped] {$(11, 1, 1)$}   (4);
\path[-latex] (2) edge node[above,sloped] {$(6, 1, 1)$}   (5);

\path[-latex] (3) edge node[below,sloped] {$(9, 1, 1)$}   (4); 

\path[-latex] (4) edge node[below,sloped] {$(-8, 1, 1)$}   (5);
\path[-latex] (4) edge node[below,sloped] {$(13, 2, 1)$}   (6); 

\path[-latex] (5) edge node[above,sloped] {$(9, 1, 1)$}   (6);

\end{tikzpicture}
\caption{Graph with fuzzy weights z \cite{id:Hernandes2007Shortest}.}
\label{fig:fuzzy-graph-2}
\end{figure}

The second example is also taken from \cite{id:Hernandes2007Shortest}.
The graph is presented in Fig. \ref{fig:fuzzy-graph-2}.
Tab. \ref{tab:example:1} presents the paths taken from \cite{id:Hernandes2007Shortest}. We have added one more column (EFN) with the results elaborated with \efn s. 
Bellow, we present non-direct paths elaborated for the threshold
$\xi = 0.3$:
\begin{itemize}
\item The path from node 1 to 3 to has length $(6, 3, 5)$. The path is  $1 \xrightarrow{(2, 1, 1)} 2 \xrightarrow{(4, 3, 5)]} 3$.
\item The path from node 1 to 4 to has length $(13, 1, 1)$. The path is $1 \xrightarrow{(2, 1, 1)} 2 \xrightarrow{(11, 1, 1)]} 4$.
\item The path from node 1 to 5 to has length $(5, 1, 1)$. The path is $1 \xrightarrow{(2, 1, 1)} 2 \xrightarrow{(11, 1, 1)} 4 \xrightarrow{(-8, 1, 1)]} 5$.
\item The path from node 1 to 6 to has length $(14, 1, 1)$. The path is $1 \xrightarrow{(2, 1, 1)} 2 \xrightarrow{(11, 1, 1)} 4 \xrightarrow{(-8, 1, 1)} 5 \xrightarrow{(9, 1, 1)} 6$.
\end{itemize}

In this example, the Okada-Soper approach elaborates longer paths than other techniques.

\begin{table}
\centering
\caption{The order relations for Example 2 are taken from \cite{id:Hernandes2007Shortest}.  Yager (Y),  Liou-Wang (LW),  García-Lamata (GL), Okada-Soper (OS),  Nayeem-Pal (NP),  Dubois-Prade (DP). The column \efn\ presents path length elaborated with our approach.}
\label{tab:example:2}
\begin{tabular}{rlllp{3cm}}
	\toprule
	                                           end & shortest path            & \multicolumn{2}{c}{path cost} & order relation \\
	\cmidrule(lr){3-4} 
 node &                          & \cite{id:Hernandes2007Shortest} &  EFN                           & \\
	\midrule
  2 & $1 \to 2$                          & $( 2, 1, 1)$  & $(2, 1, 1)$  & all\\
  3 & $1 \to 3$                          & $( 7, 2, 2)$  & n/a  & OS ($\varepsilon = 0$ and 0.5), LW ($\lambda = 1$), GL ($\lambda=1, \delta=0.5$)\\
  3 & $1 \to 2 \to 3$                    & $( 6, 4, 6)$  & $(6, 3, 5)$  & all, except LW ($\lambda=1$), GL ($\lambda=1, \delta=0$)\\
  4 & $1 \to 2 \to 4  $                  & $(13, 2, 2)$  & $(13, 1, 1)$  & all\\
  4 & $1 \to 2 \to 3 \to 4$              & $(15, 5, 7)$  & n/a  & OS $(\varepsilon=0)$ \\
  5 & $1 \to 2 \to 4 \to 5$              & $( 5, 3, 3)$  & $(5, 1, 1)$  & all\\
  5 & $1 \to 2 \to 3 \to 4 \to 5$        & $( 7, 6, 8)$  & n/a  & OS $(\varepsilon=0)$ \\
  6 & $1 \to 2 \to 4 \to 5 \to 6$        & $(14, 4, 4)$  & $(14, 1, 1)$  & all\\
  6 & $1 \to 2 \to 3 \to 4 \to 5 \to 6$  & $(16, 7, 9)$  & n/a  & OS $(\varepsilon=0)$ \\
\bottomrule	       
\end{tabular}
\end{table}

\subsubsection{Example 3}
In examples presented in Sec. \ref{sec:ex:1} and \ref{sec:ex:2}, our approach returns unique paths. This is because the fuzziness of edges not fuzzy enough to influence the ordering of fuzzy numbers.

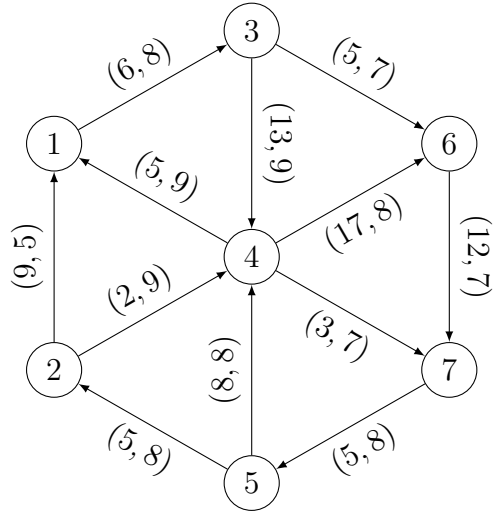
\begin{figure}
\centering 
\begin{tikzpicture}

\draw (-2.6,  1.5) node (1) [circle,draw=black] {1};
\draw (-2.6, -1.5) node (2) [circle,draw=black] {2};

\draw (0, 3) node (3) [circle,draw=black] {3};
\draw (0, 0) node (4) [circle,draw=black] {4};
\draw (0,-3) node (5) [circle,draw=black] {5};

\draw (2.6,  1.5) node (6) [circle,draw=black] {6};
\draw (2.6, -1.5) node (7) [circle,draw=black] {7};

\path[-latex] (1) edge node[above, sloped] {$(6,8)$}  (3);
\path[-latex] (2) edge node[above, sloped] {$(9,5)$}  (1);
\path[-latex] (2) edge node[above, sloped] {$(2,9)$}  (4);
\path[-latex] (3) edge node[above, sloped] {$(13,9)$} (4);
\path[-latex] (3) edge node[above, sloped] {$(5,7)$}  (6);
\path[-latex] (4) edge node[above, sloped] {$(5,9)$}  (1);
\path[-latex] (4) edge node[below, sloped] {$(17,8)$}  (6);
\path[-latex] (4) edge node[below, sloped] {$(3,7)$}  (7);
\path[-latex] (5) edge node[below,sloped] {$(5,8)$}  (2);
\path[-latex] (5) edge node[above, sloped] {$(8,8)$}  (4);
\path[-latex] (6) edge node[above, sloped] {$(12,7)$}  (7);
\path[-latex] (7) edge node[below,sloped] {$(5,8)$}  (5);

\end{tikzpicture}
\caption{Graph with fuzzy weights.}
\label{fig:graph-fuzzy-3}
\end{figure}

In this example, we would like to show the influence of fuzziness on the elaborated paths. Here we use symmetric triangular \efn s. Fuzzy numbers are encoded as a pair of values: a core and a half of support width.
The graph is presented in Fig. \ref{fig:graph-fuzzy-3}.

The fuzziness of fuzzy numbers is significant. The thresholding enables various influence of fuzziness on the final result.
For the threshold $\xi = 0.3$:
\begin{itemize}
\item The path from node 2 to 1 has length $(9, 5)$. The path is $2 \xrightarrow{(9, 5)} 1$.
\item The path from node 7 to 6 has length $(30, 8)$. The path is $7 \xrightarrow{(5, 8)} 5 \xrightarrow{(8, 8)} 4 \xrightarrow{(17, 8)}6$.
\end{itemize}
The paths for these pairs of nodes for $\xi = 0.1$ are different:
\begin{itemize}
\item The path from node 2 to 1 has length $(7, 9)$. The path is $2 \xrightarrow{(2, 9)} 4 \xrightarrow{(5, 9)} 1$.
\item The path from node 7 to 6 has length $(28, 9)$. The path is $7 \xrightarrow{(5, 8)} 5 \xrightarrow{(5, 8)} 2 \xrightarrow{(2, 9)} 4 \xrightarrow{(5, 9)} 1 \xrightarrow{(6, 8)} 3 \xrightarrow{(5, 7)} 6$.
\end{itemize}

\section{Conclusions}
The paper presents an application of extended fuzzy numbers (\efn s) to the problem of ordering of fuzzy numbers.
The \efn s with operations and relational operations enable to operate on fuzzy terms directly without any need of defuzzification, integration or any iterative procedures.
The construction of \efn s is simple and operations on \efn s are computationally efficient. 
Multiple types of \efn s can be constructed with T-norms what makes the number of types of \efn s infinite.
The application examples presented in the paper show that \efn s can be sucessfully and efficiently applied instead of real numbers in problems where uncertainty and imprecision are present.
The C++ implementation of \efn s is available from the free public GitHub repository: \url{github.com/ksiminski/ksi-extensional-fuzzy-numbers}


\begin{thebibliography}{10}

\bibitem{id:Behounek2026Maxima}
Libor Běhounek.
\newblock Maxima and minima in fuzzified linear orderings.
\newblock {\em Fuzzy Sets and Systems}, 289:82--93, 2016.
\newblock Theme: Non-additive Measures and Algebra.

\bibitem{id:Cheng2004Group}
Chi-Bin Cheng.
\newblock Group opinion aggregation based on a grading process: a method for
  constructing triangular fuzzy numbers.
\newblock {\em International Journal Computers and Mathematics with
  Applications}, 48:1619--1632, 2004.

\bibitem{id:Ciucci2016Orthopairs}
Davide Ciucci.
\newblock Orthopairs and granular computing.
\newblock {\em Granular Computing}, 1:159--170, 2016.

\bibitem{id:Dubois1978Operations}
D.~Dubois and H.~Prade.
\newblock Operations on fuzzy numbers.
\newblock {\em International Journal of Systems Science}, 9(6):613--626, 1978.

\bibitem{id:Dubois1983Ranking}
Didier Dubois and Henri Prade.
\newblock Ranking fuzzy numbers in the setting of possibility theory.
\newblock {\em Information Sciences}, 30(3):183--224, 1983.

\bibitem{id:Garcia2005Fuzzy}
M.~Socorro García-Cascales and Maria Lamata.
\newblock The fuzzy sets in maintenance process.
\newblock In {\em Proceedings of the Joint 4th Conference of the European
  Society for Fuzzy Logic and Technology and the 11th Rencontres Francophones
  sur la Logique Floue et ses Applications}, pages 118--123, 01 2005.

\bibitem{id:Garcia-Zamora2024Admissible}
Diego García-Zamora, Anderson Cruz, Fernando Neres, Regivan~H.N. Santiago,
  Antonio~F. {Roldán López de Hierro}, Rui Paiva, Graçaliz.~P. Dimuro, Luis
  Martínez, Benjamín Bedregal, and Humberto Bustince.
\newblock Admissible owa operators for fuzzy numbers.
\newblock {\em Fuzzy Sets and Systems}, 480:108863, 2024.

\bibitem{id:GeramiSeresht2019Computational}
Nima {Gerami Seresht} and Aminah~Robinson Fayek.
\newblock Computational method for fuzzy arithmetic operations on triangular
  fuzzy numbers by extension principle.
\newblock {\em International Journal of Approximate Reasoning}, 106:172--193,
  2019.

\bibitem{id:Hassanzadeh2013Genetic}
Reza Hassanzadeh, Iraj Mahdavi, Nezam Mahdavi-Amiri, and Ali Tajdin.
\newblock A genetic algorithm for solving fuzzy shortest path problems with
  mixed fuzzy arc lengths.
\newblock {\em Mathematical and Computer Modelling}, 57(1):84--99, 2013.
\newblock Mathematical and Computer Modelling in Power Control and
  Optimization.

\bibitem{id:Hernandes2007Shortest}
Fábio Hernandes, Maria~Teresa Lamata, José~Luis Verdegay, and Akebo Yamakami.
\newblock The shortest path problem on networks with fuzzy parameters.
\newblock {\em Fuzzy Sets and Systems}, 158(14):1561--1570, 2007.

\bibitem{id:Holcapek2012Arithmethics-I}
Michal Holčapek and Martin Štěpnička.
\newblock Arithmetics of extensional fuzzy numbers - part {I}: Introduction.
\newblock In {\em 2012 {IEEE} International Conference on Fuzzy Systems}, pages
  1--8, 2012.

\bibitem{id:Holcapek2012Arithmethics-II}
Michal Holčapek and Martin Štěpnička.
\newblock Arithmetics of extensional fuzzy numbers - part {II}: Algebraic
  framework.
\newblock In {\em 2012 IEEE International Conference on Fuzzy Systems}, pages
  1--8, 2012.

\bibitem{id:Holcapek2014MI-Algebras}
Michal Holčapek and Martin Štěpnička.
\newblock {MI}-algebras: A new framework for arithmetics of (extensional) fuzzy
  numbers.
\newblock {\em Fuzzy Sets and Systems}, 257:102--131, 2014.
\newblock Special Issue on Fuzzy Numbers and Their Applications.

\bibitem{id:Kuchta2002Generalisation}
Dorota Kuchta.
\newblock A generalisation of an algorithm solving the fuzzy multiple choice
  knapsack problem.
\newblock {\em Fuzzy Sets and Systems}, 127(2):131--140, 2002.

\bibitem{id:Lin2021Genetic}
Lihua Lin, Chuzheng Wu, and Li~Ma.
\newblock A genetic algorithm for the fuzzy shortest path problem in a fuzzy
  network.
\newblock {\em Complex and Intelligent Systems}, 7:225--234, 2021.

\bibitem{id:Liou1992Ranking}
Tian-Shy Liou and Mao-Jiun~J. Wang.
\newblock Ranking fuzzy numbers with integral value.
\newblock {\em Fuzzy Sets and Systems}, 50(3):247--255, 1992.

\bibitem{id:Ma1999New}
Ming Ma, Menahem Friedman, and Abraham Kandel.
\newblock A new fuzzy arithmetic.
\newblock {\em Fuzzy Sets and Systems}, 108(1):83--90, 1999.

\bibitem{id:Mahdavi2009Dynamic}
Iraj Mahdavi, Rahele Nourifar, Armaghan Heidarzade, and Nezam~Mahdavi Amiri.
\newblock A dynamic programming approach for finding shortest chains in a fuzzy
  network.
\newblock {\em Applied Soft Computing}, 9(2):503--511, 2009.

\bibitem{id:Mukherjee2023Brief}
Asesh~Kumar Mukherjee, Kamal~Hossain Gazi, Soheil Salahshour, Arijit Ghosh, and
  Sankar~Prasad Mondal.
\newblock A brief analysis and interpretation on arithmetic operations of fuzzy
  numbers.
\newblock {\em Results in Control and Optimization}, 13:100312, 2023.

\bibitem{id:Nayeem2005Shortest}
S.M.A. Nayeem and M.~Pal.
\newblock Shortest path problem on a network with imprecise edge weight.
\newblock {\em Fuzzy Optimization and Decision Making}, 4:293--312, 2005.

\bibitem{id:Nguyen1978Note}
Hung~T Nguyen.
\newblock A note on the extension principle for fuzzy sets.
\newblock {\em Journal of Mathematical Analysis and Applications},
  64(2):369--380, 1978.

\bibitem{id:Okada2000Shortest}
Shinkoh Okada and Timothy Soper.
\newblock A shortest path problem on a network with fuzzy arc lengths.
\newblock {\em Fuzzy Sets and Systems}, 109(1):129--140, 2000.

\bibitem{id:Pedrycz2012GranularRepresentation}
Adam Pedrycz, Kaoru Hirota, Witold Pedrycz, and Fangyan Dong.
\newblock Granular representation and granular computing with fuzzy sets.
\newblock {\em Fuzzy Sets and Systems}, 203:17--32, 2012.

\bibitem{id:Pedrycz2013Granular}
Witold Pedrycz.
\newblock {\em Granular Computing: Analysis and Design of Intelligent Systems}.
\newblock CRC Press, 2013.

\bibitem{id:Pedrycz2013Building}
Witold Pedrycz and Wladyslaw Homenda.
\newblock Building the fundamentals of granular computing: A principle of
  justifiable granularity.
\newblock {\em Applied Soft Computing}, 13(10):4209--4218, 2013.

\bibitem{id:Pedrycz2015Data}
Witold Pedrycz, Giancarlo Succi, Alberto Sillitti, and Joana Iljazi.
\newblock Data description: A general framework of information granules.
\newblock {\em Knowledge-Based Systems}, 80:98--108, 2015.

\bibitem{id:Piegat2015Computing}
Andrzej Piegat and Marcin Pluciński.
\newblock Computing with words with the use of inverse rdm models of membership
  functions.
\newblock {\em International Journal of Applied Mathematics and Computer
  Science}, 25(3):675--688, 2015.

\bibitem{id:Pieta2021Applications}
Piotr Pieta and Tomasz Szmuc.
\newblock Applications of rough sets in big data analysis: An overview.
\newblock {\em International Journal of Applied Mathematics and Computer
  Science}, 31(4):659--683, 2021.

\bibitem{id:Salehi2015Systematic}
Saber Salehi, Ali Selamat, and Hamido Fujita.
\newblock Systematic mapping study on granular computing.
\newblock {\em Knowledge-Based Systems}, 80:78--97, 2015.

\bibitem{id:Shaocheng1994Interval}
Tong Shaocheng.
\newblock Interval number and fuzzy number linear programmings.
\newblock {\em Fuzzy Sets and Systems}, 66(3):301--306, 1994.

\bibitem{id:Shifei2010Research}
D.~Shifei, X.~Li, Z.~Hong, and Z.~Liwen.
\newblock Research and progress of cluster algorithms based on granular
  computing.
\newblock {\em International Journal of Digital Content Technology and its
  Applications}, 4(5):96--104, 2010.

\bibitem{id:Siminski2020GrFCM}
Krzysztof Siminski.
\newblock {GrFCM} - granular clustering of granular data.
\newblock In Aleksandra Gruca, Tadeusz Czachórski, Sebastian Deorowicz,
  Katarzyna Harezlak, and Agnieszka Piotrowska, editors, {\em Man-Machine
  Interactions 6}, pages 111--121, Cham, 2020. Springer International
  Publishing.

\bibitem{id:Siminski2021GrNFS}
Krzysztof Siminski.
\newblock {GrNFS} - granular neuro-fuzzy system for regression in large volume
  data.
\newblock {\em International Journal of Applied Mathematics and Computer
  Science}, 31(3):445--459, 2021.

\bibitem{id:Siminski2021Outlier}
Krzysztof Siminski.
\newblock An outlier-robust neuro-fuzzy system for classification and
  regression.
\newblock {\em International Journal of Applied Mathematics and Computer
  Science}, 31(2):303--319, 2021.

\bibitem{id:Siminski2022Prototype}
Krzysztof Siminski.
\newblock Prototype based granular neuro-fuzzy system for regression task.
\newblock {\em Fuzzy Sets and Systems}, 449:56--78, 2022.

\bibitem{id:Siminski20233WDNFS}
Krzysztof Siminski.
\newblock {3WDNFS} - three-way decision neuro-fuzzy system for classification.
\newblock {\em Fuzzy Sets and Systems}, 466:108432, 2023.

\bibitem{id:Skowron2016Interactive}
Andrzej Skowron, Andrzej Jankowski, and Soma Dutta.
\newblock Interactive granular computing.
\newblock {\em Granular Computing}, 1:95--113, 2016.

\bibitem{id:Stepnicka2020Arithmetics}
Martin Štěpnička, Nicole Škorupová, and Michal Holčapek.
\newblock From arithmetics of extensional fuzzy numbers to their distances.
\newblock In {\em 2020 IEEE International Conference on Fuzzy Systems
  (FUZZ-IEEE)}, pages 1--8, 2020.

\bibitem{id:Stepnicka2020Properties}
Martin Štěpnička, Nicole Škorupová, and Michal Holčapek.
\newblock On the properties of orderings of extensional fuzzy numbers.
\newblock In {\em 2020 IEEE International Conference on Fuzzy Systems
  (FUZZ-IEEE)}, pages 1--7, 2020.

\bibitem{id:Suchy2023GrDBSCAN}
Dawid Suchy and Krzysztof Siminski.
\newblock Grdbscan: A granular density-based clustering algorithm.
\newblock {\em International Journal of Applied Mathematics and Computer
  Science}, 33(2):297--312, 2023.

\bibitem{id:Szpilrajn1930Extension}
Edward Szpilrajn.
\newblock Sur l'extension de l'ordre partiel.
\newblock {\em Fundamenta Mathematicae}, 16:386--389, 1930.

\bibitem{id:Tajdin2010Computing}
Ali Tajdin, Iraj Mahdavi, Nezam Mahdavi-Amiri, and Bahram Sadeghpour-Gildeh.
\newblock Computing a fuzzy shortest path in a network with mixed fuzzy arc
  lengths using $\alpha$-cuts.
\newblock {\em Computers and Mathematics with Applications}, 60(4):989--1002,
  2010.

\bibitem{id:Tolga2020Finite}
A.~Cagri Tolga, I.~Burak Parlak, and Oscar Castillo.
\newblock Finite-interval-valued type-2 gaussian fuzzy numbers applied to fuzzy
  todim in a healthcare problem.
\newblock {\em Engineering Applications of Artificial Intelligence}, 87:103352,
  2020.

\bibitem{id:Wang2021Designing}
Dan Wang, Peng Nie, Xiubin Zhu, Witold Pedrycz, and Zhiwu Li.
\newblock Designing of higher order information granules through clustering
  heterogeneous granular data.
\newblock {\em Applied Soft Computing}, 112:107820, 2021.

\bibitem{id:Xia2022New}
Sisi Xia, Lin Chen, Siya Liu, and Haoran Yang.
\newblock A new method for decision making problems with redundant and
  incomplete information based on incomplete soft sets: from crisp to fuzzy.
\newblock {\em International Journal of Applied Mathematics and Computer
  Science}, 32(4):657--669, 2022.

\bibitem{id:Yager1978Ranking}
Ronald~R. Yager.
\newblock Ranking fuzzy subsets over the unit interval.
\newblock In {\em 1978 IEEE Conference on Decision and Control including the
  17th Symposium on Adaptive Processes}, pages 1435--1437, 1978.

\bibitem{id:Yager1980Choosing}
Ronald~R. Yager.
\newblock On choosing between fuzzy subsets.
\newblock {\em Kybernetes}, 9(2):151--154, 1980.

\bibitem{id:Yager1981Procedure}
Ronald~R. Yager.
\newblock A procedure for ordering fuzzy subsets of the unit interval.
\newblock {\em Information Sciences}, 24:143--161, 1981.

\bibitem{id:Yager1988Ordered}
Ronald~R. Yager.
\newblock On ordered weighted averaging aggregation operators in multicriteria
  decisionmaking.
\newblock {\em {IEEE} Transactions on Systems, Man, and Cybernetics},
  18(1):183--190, 1988.

\bibitem{id:Yao2013Granular}
J.~T. Yao, A.~V. Vasilakos, and W.~Pedrycz.
\newblock Granular computing: Perspectives and challenges.
\newblock {\em {IEEE} Transactions on Cybernetics}, 43(6):1977--1989, Dec 2013.

\bibitem{id:Yao2007Granular}
Y.~Yao and N.~Zhong.
\newblock Granular computing.
\newblock In B.W. Wah, editor, {\em Wiley Encyclopedia of Computer Science and
  Engineering}. Wiley, 2007.

\bibitem{id:Yao2007Art}
Yiyu Yao.
\newblock The art of granular computing.
\newblock In Marzena Kryszkiewicz, James~F. Peters, Henryk Rybinski, and
  Andrzej Skowron, editors, {\em Rough Sets and Intelligent Systems Paradigms},
  pages 101--112, Berlin, Heidelberg, 2007. Springer Berlin Heidelberg.

\bibitem{id:Yao2008GranularPast}
Yiyu Yao.
\newblock Granular computing: Past, present and future.
\newblock In {\em The 2008 {IEEE} International Conference on Granular
  Computing, GrC 2008, Hangzhou, China, 26-28 August 2008}, pages 80--85, 2008.

\bibitem{id:Yao2009Three}
Yiyu Yao.
\newblock Three-way decision: An interpretation of rules in rough set theory.
\newblock In Peng Wen, Yuefeng Li, Lech Polkowski, Yiyu Yao, Shusaku Tsumoto,
  and Guoyin Wang, editors, {\em Rough Sets and Knowledge Technology}, pages
  642--649, Berlin, Heidelberg, 2009. Springer Berlin Heidelberg.

\bibitem{id:Yao2011Superiority}
Yiyu Yao.
\newblock The superiority of three-way decisions in probabilistic rough set
  models.
\newblock {\em Information Sciences}, 181(6):1080--1096, 2011.

\bibitem{id:Yao2016triarchic}
Yiyu Yao.
\newblock A triarchic theory of granular computing.
\newblock {\em Granular Computing}, 1:145--157, 2016.

\bibitem{id:Yao2018Three}
Yiyu Yao.
\newblock Three-way decision and granular computing.
\newblock {\em International Journal of Approximate Reasoning}, 103:107--123,
  2018.

\bibitem{id:Yao2020Three-way}
Yiyu Yao.
\newblock Three-way granular computing, rough sets, and formal concept
  analysis.
\newblock {\em International Journal of Approximate Reasoning}, 116:106--125,
  2020.

\bibitem{id:Zadeh1979Fuzzy}
Lotfi~A. Zadeh.
\newblock Fuzzy sets and information granularity.
\newblock In N.~Gupta, R.~Ragade, and R.~Yager, editors, {\em Advances in Fuzzy
  Set Theory and Applications}, pages 3--18. North-Holland Publishing Co.,
  1979.

\bibitem{id:Zadeh1997Toward}
Lotfi~A. Zadeh.
\newblock Toward a theory of fuzzy information granulation and its centrality
  in human reasoning and fuzzy logic.
\newblock {\em Fuzzy Sets and Systems}, 90(2):111--127, 1997.

\bibitem{id:Zadeh2002Computing}
Lotfi~A. Zadeh.
\newblock From computing with numbers to computing with words—from
  manipulation of measurements to manipulation of perceptions.
\newblock {\em International Journal of Applied Mathematics and Computer
  Science}, 12(3):307--324, 2002.

\bibitem{id:Zumelzu2022Admissible}
Nicolás Zumelzu, Benjamín Bedregal, Edmundo Mansilla, Humberto Bustince, and
  Roberto Díaz.
\newblock Admissible orders on fuzzy numbers.
\newblock {\em IEEE Transactions on Fuzzy Systems}, 30(11):4788--4799, 2022.
\end{thebibliography}
\end{document}